\documentclass[10pt,journal,compsoc]{IEEEtran}

\ifCLASSOPTIONcompsoc
\usepackage[nocompress]{cite}
\else
\usepackage{cite}
\fi

\usepackage{amsmath,amsfonts}
\usepackage{algorithmic}
\usepackage{algorithm}
\usepackage{array}
\usepackage{textcomp}
\usepackage{stfloats}
\usepackage{url}
\usepackage{verbatim}
\usepackage{graphicx}
\usepackage{fancyhdr,lipsum}
\renewcommand{\headrulewidth}{0pt}
\renewcommand{\footrulewidth}{0pt}
\usepackage{subfigure}
\usepackage{caption}
\usepackage{bbm}
\usepackage{dsfont}
\usepackage{setspace}
\usepackage{multirow}
\usepackage{tabularx}
\usepackage{booktabs}

\usepackage{color}
\usepackage[table]{xcolor}
\usepackage{mdframed}

\DeclareMathOperator*{\argmin}{\arg\min}
\DeclareMathOperator*{\argmax}{\arg\max}

\newcommand{\cX}{\mathcal{X}}

\newcommand{\cH}{\mathcal{H}}
\newcommand{\cL}{\ell}
\newcommand{\cF}{\mathcal{F}}

\newcommand{\bR}{\mathbb{R}}

\newcommand{\ST}{\mathrm{s.t.}}
\newcommand{\cW}{W}
\newcommand{\cU}{\mathcal{U}}

\begin{document}
\title{BadLabel: A Robust Perspective on \\ Evaluating and Enhancing Label-noise Learning}

\author{Jingfeng Zhang$^{1,2*}$, Bo Song$^{3*}$, \\ Haohan Wang$^{4}$, Bo Han$^{5,2}$, Tongliang Liu$^{6,2}$, Lei Liu$^{3,7}$, and Masashi Sugiyama$^{2,8}$
\thanks{The first two authors made equal contributions.}
\thanks{$^1$the University of Auckland, $^2$RIKEN AIP, $^3$Shandong University, $^4$the University of Illinois Urbana-Champaign, $^5$Hong Kong Baptist University, $^6$the University of Sydney, $^7$Shandong Research Institute of Industrial Technology, $^8$the University of Tokyo \\Correspond to jingfeng.zhang@auckland.ac.nz}

}

\markboth{IEEE T-PAMI~2024}%
{Shell \MakeLowercase{\textit{et al.}}: A Sample Article Using IEEEtran.cls for IEEE Journals}


\IEEEtitleabstractindextext{%
\begin{abstract}
	Label-noise learning (LNL) aims to increase the model's generalization given training data with noisy labels.
	To facilitate practical LNL algorithms, researchers have proposed different label noise types, ranging from class-conditional to instance-dependent noises. 
	In this paper, we introduce a novel label noise type called \textit{BadLabel}, which can significantly degrade the performance of existing LNL algorithms by a large margin.
	BadLabel is crafted based on the label-flipping attack against standard classification, where specific samples are selected and their labels are flipped to other labels so that the loss values of clean and noisy labels become indistinguishable.
	To address the challenge posed by BadLabel, we further propose a robust LNL method that perturbs the labels in an adversarial manner at each epoch to make the loss values of clean and noisy labels again distinguishable. 
	Once we select a small set of (mostly) clean labeled data, we can apply the techniques of semi-supervised learning to train the model accurately. 
	Empirically, our experimental results demonstrate that existing LNL algorithms are vulnerable to the newly introduced BadLabel noise type, while our proposed robust LNL method can effectively improve the generalization performance of the model under various types of label noise.
	The new dataset of noisy labels and the source codes of robust LNL algorithms are available at https://github.com/zjfheart/BadLabels.
	%
\end{abstract}

}

\maketitle

\section{Introduction}
\IEEEPARstart{L}{abel-noise} learning (LNL) has become increasingly important in deep learning classification problems due to the high cost and often inaccuracy of annotations in large-scale datasets~\cite{yan2014learning,li2017webvision,yu2018learning,natarajan2013learning}. To facilitate the development of effective LNL algorithms, researchers have designed various noise types ranging from class-conditional noise (such as symmetry-flipping~\cite{angluin1988learning, natarajan2013learning, van2015learning} and pair-flipping~\cite{han2018co} noise) to instance-dependent noise~\cite{xia2020part, zhang2020learning, chen2021beyond}. In class-conditional noise, a data point's label has a fixed probability of being flipped to another label; whereas in instance-dependent noise, the label-flipping probability depends on both the true label and features of each data point. 


However, it remains unclear whether the existing LNL algorithms, such as DivideMix~\cite{li2019dividemix}, SOP~\cite{liu22sop}, and ProMix~\cite{wang2022promix}, are capable of handling even more challenging types of label noise. In high-stake applications such as medicine~\cite{ju2022improving,karimi2020deep} and cybersecurity~\cite{taheri2020defending,catal2022applications}, where the use of machine learning techniques is under close scrutiny, it is crucial for practitioners to be aware of the limitations of existing LNL algorithms and to employ the most robust LNL methods to ensure accurate models under various types of label noise. 
This motivates the development of challenging label noise types that can expose the vulnerabilities of existing LNL algorithms. 
Furthermore, the challenging noises can facilitate the development of more robust LNL algorithms that are applicable not only to common but also to rare types of noise.


To this end, we introduce a new type of label noise called BadLabel, which is created using label-flipping attacks~\cite{biggio2011support,zhao2017efficient} on a standard multi-class classification task~\cite{aly2005survey}. 
The discrete label space poses a challenge as it hinders the direct optimization of labels for maximizing a loss. To overcome this issue, we propose a surrogate \emph{flag} array that can produce a label-flipping strategy. 
As shown in Algorithm~\ref{alg:BadLabel}, we optimize the flag over several training epochs, and the flag array in the end determines which data to flip its label and how the label should be flipped. 
This approach enables us to effectively handle the discrete label space and generate a challenging label noise that can expose the vulnerabilities of existing label noise learning algorithms.


\begin{figure*}[t]
	\vspace{-2mm}
	\centering
 \includegraphics[scale=0.54]{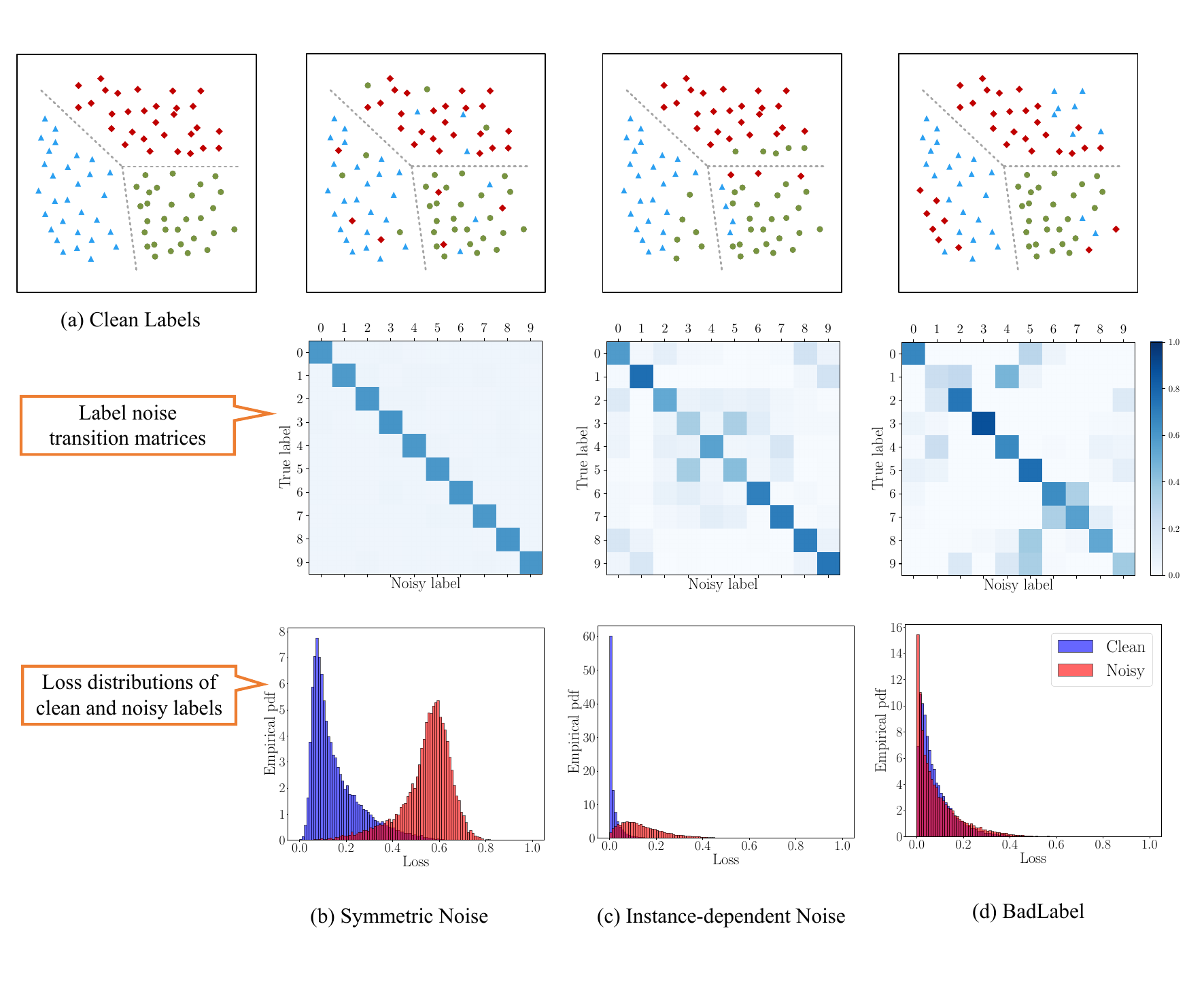}
 \vspace{-10mm}
	\caption{Comparison of different types of label noise: (a) Clean labels, representing a noise-free dataset. (b) Symmetric noise, where the label noise is distributed randomly in each class. (c) Instance-dependent noise, where the label noise is concentrated near the class boundaries. (d) BadLabel, where the label noise is far from the class boundaries.
 \textbf{Top row}: Synthetic three-class examples.  
 \textbf{Middle row}: Empirical transition matrices of different types of label noise on the CIFAR-10 dataset. 
 \textbf{Bottom row}: Loss distributions of clean and noisy labels of the CIFAR-10 dataset, given a properly trained model.} 
	\label{fig:prob_matrix_and_loss_dist}
	\vspace{-4mm}
\end{figure*}

We visualized and compared different types of label noise in Figure~\ref{fig:prob_matrix_and_loss_dist}, with a focus on the challenging nature of the BadLabel to LNL algorithms.
In the top row, we used a synthetic three-class classification to show that BadLabel often flips the labels of samples that are located far from the class boundary, leading to clusters of noisy-label data. These clusters can significantly mislead the conventional learning algorithm and result in learning wrong decision boundaries or failure in training.
In the middle and bottom rows, we used the real-world CIFAR-10 dataset~\cite{krizhevsky2009learning} to compare different label-flipping strategies. In the middle row, we visualized the empirical transition matrices of label noise, demonstrating that BadLabel has distinct corruptions from other types of label noise~\cite{han2018masking}. In the bottom row, we used a well-performing classifier to compute the loss values of clean and noisy labels over the whole training set and plot the loss distribution. We observed that BadLabel makes the noisy labels less distinguishable from the clean labels in terms of loss values compared to other noise types. It is worth noting that the loss value is a crucial metric in LNL algorithms to select and correct samples~\cite{han2018co}. Therefore, BadLabel is indeed a challenging type of label noise to the existing LNL algorithms.


To deal with BadLabel, we propose a robust LNL algorithm called Robust DivideMix. The standard DivideMix~\cite{li2019dividemix} models the loss distribution with a Gaussian Mixture Model (GMM)~\cite{permuter2006study} to divide the training data into a clean labeled set and an unlabeled set, and then applies a semi-supervised learning technique such as MixMatch~\cite{berthelot2019mixmatch}. 
However, the GMM fails to model BadLabel effectively because the loss values of noisy and clean labels are not always distinguishable, as illustrated in the bottom raw of Figure 1. 
To address this issue, our Robust DivideMix perturbs labels in an adversarial manner to aid in selecting and splitting clean and noisy labels. We then apply the BayesGMM~\cite{RobertsHRP1998bayesian} and MixMatch to divide the data and train the models, enabling our method to handle various types of label noise, including BadLabel.


Our contributions can be summarized as follows. 
\begin{itemize}
    \vspace{-2mm}
	\item We are the first to introduce a challenging type of label noise, BadLabel (Algorithm~\ref{alg:BadLabel}). 
	We mathematically analyze and justify our proposed algorithm that can reasonably produce a BadLabel dataset (see Section~\ref{Sec:method_badlabel}).  
	\item We demonstrate that BadLabel noise can significantly degrade the performance of 11 state-of-the-art LNL algorithms (see Section~\ref{Sec:exp-BadLabel}).
	\item We propose a robust LNL algorithm to deal with BadLabel (see Section~\ref{Sec:method_robust_dividemix}). Compared to the existing LNL algorithms, our Robust DivideMix can effectively improve the model's generalization under BadLabel. Furthermore, Robust DivideMix can maintain comparable performance with DivideMix under other types of label noise, such as symmetric and instance-dependent noises (see Section~\ref{Sec:exp-robust-dividemix}).
\end{itemize}

\section{Related Works}
We review LNL algorithms and label-flipping attacks.
\vspace{-4mm}
\subsection{Label-noise Learning (LNL) Algorithms}
Deep neural networks (DNNs) can easily memorize and overfit noisy labels and produce suboptimal models~\cite{arpit2017closer}. Many LNL algorithms have been proposed to improve the model's generalization under label noise, which can be broadly classified into four categories: module-based methods, loss-based methods, label correction methods, and sample selection methods.

Module-based methods modify the neural network modules to be more robust against label noise.
For example, Sukhbaatar et al. (2015)~\cite{sukhbaatar2015training} proposed a method called ``Hard Attention'' to train a DNN with a binary mask to filter out noisy samples.
Goldberger and Ben-Reuven (2017)~\cite{goldberger2017training} proposed a method called ``Denoising Autoencoder'' to reconstruct the clean samples from the noisy samples. Han et al. (2018)~\cite{han2018masking} proposed a method called ``Masking'' to train a neural network with a binary mask to filter out noisy labels. 

Loss-based methods design the loss function to be more robust against label noise. For example, Zhang and Sabuncu (2018)~\cite{zhang2018generalized} proposed a method called ``Generalized Cross Entropy'' to reduce the impact of noisy labels on the loss function. Liu et al. (2020)~\cite{liu2020peer} proposed a method called ``PeerLoss'' to leverage the consistency between clean and noisy labels to reduce the effect of label noise. Adversarial training~\cite{goodfellow2015exp} and label smoothing~\cite{lukasik2020does, wei2021understanding} are also commonly used techniques to regularize the loss function against label noise.

Label correction methods adjust the loss value or correct the labels to mitigate the impact of label noise. For example, Patrini et al. (2017)~\cite{patrini2017making} proposed a method called ``Forward and Backward Loss Correction'' to adjust the loss value based on the label confidence. Arazo et al. (2019)~\cite{arazo2019unsupervised} proposed a method called ``Unsupervised Data Cleaning'' to correct the noisy labels by clustering the samples. Chen et al. (2021)~\cite{chen2021beyond} proposed a method called ``Beyond Learning to Correct'' to correct the noisy labels by learning the label transition matrix.

Sample selection methods select the clean samples from the noisy dataset to improve the model's generalization performance. For example, Han et al. (2018)~\cite{han2018co} proposed a method called ``Co-teaching'' to train two neural networks on different subsets of the dataset and let each network select the clean samples for the other network. Li et al. (2019)~\cite{li2019dividemix} proposed a method called ``DivideMix'' to divide the dataset into a labeled set with clean samples and an unlabeled set, and then applied the semi-supervised learning. Wang et al. (2022)~\cite{wang2022promix} proposed a method called ``ProMix'' to select the clean samples based on their similarity to the noisy samples.

The existing LNL algorithms have achieved excellent performance under the conventional types of label noise such as symmetry-flipping noise and instance-dependent noise. However, the performance of LNL algorithms under more challenging types of label noise, such as BadLabel, is still an open research problem. BadLabel refers to the label noise that intentionally flips the labels of samples. The noisy-label data are located far from the decision boundary, leading to the clusters, which can significantly mislead the conventional learning algorithms and result in learning the wrong decision boundaries or failure in training. This paper proposes the first attempt to handle BadLabel, but more research is needed to develop robust LNL algorithms under this type of label noise.

\subsection{Label-flipping Attacks}
The earliest label-flipping attacks date back to bypassing the detection of spam email: Barreno et al. (2010)~\cite{barreno2010security} purposely gave the benign emails the ``spam'' labels. 
In the following year, Biggio et al. (2011)~\cite{biggio2011support} crafted adversarial labels against the support vector machines (SVMs)~\cite{noble2006support}. 
Xiao et al. (2012)~\cite{xiao2012adversarial} further reduced the SVMs' generalization by formulating a label-flipping optimization problem and maximizing the classification loss.
Recently, Zhao et al. (2017)~\cite{zhao2017efficient} and Paudice et al. (2018)~\cite{paudice2018label} extended label-flipping attacks to other linear classifiers. 

However, the prior arts only focused on attacking the simple linear models and binary classification tasks. In this paper, we extend the label-flipping attacks to DNNs and multi-class classification tasks, which are the settings commonly considered by those state-of-the-art LNL algorithms~\cite{li2019dividemix, liu22sop, wang2022promix}. 
We craft a challenging type of label noise via designing a label-flipping attack on the multi-classifications tasks, and propose a novel LNL algorithm that can cope with such challenging label noise.

\section{BadLabel---A Challenging Dataset}
\label{Sec:method_badlabel}
In this section, we aim to craft a challenging type of label noise---BadLabel. 
To this end, we design a label-flipping attack algorithm against a standard multi-classification task.

\subsection{Objective of BadLabel}
First, we review the learning objective of the standard multi-classification. Given a $C$-class training set $D= \big \{(x_{i}, y_{i})|x_{i} \in \bR^{d}, y_{i} \in \{0,\ldots, C-1\} \big \}_{i=1}^{n}$ where $y_{i}$ is the clean label of $x_{i}$, we can formulate the empirical learning objective as follows.
\begin{equation}
	\label{eq:dnns-objecetive}
	\mathop{\arg\min}\limits_{f \in \cF} \frac{1}{n} \sum\limits_{i=1}^{n} \cL(y_{i}, f(x_{i})), 
\end{equation}
where $f$ denotes a classifier (i.e., a DNN in this paper), $\cF$ is the hypothesis space, $\cL$ is the loss function for optimization.

Second, we design an objective function of label-flipping attack against a standard multi-class classification task. 
Given a clean training set $D$, we flip $(100\times\rho)\%$ of the clean labels that maximize the loss $\cL$, which is formulated as follows.
\begin{align}
	\label{eq:attack-objective}
	\mathop{\mathbb{E}}_{f\in \cF} \frac{1}{n} \sum\limits_{i=1}^{n} \{\max\limits_{y'_{i}} \cL(y'_{i}, f(x_{i}))\}  \nonumber \\
	\ST \;\; \frac{1}{n}  \sum\limits_{i=1}^{n} \mathbbm{1}_{\{y_{i}\}}(y'_{i}) = 1-\rho,  
\end{align}
where $y'_{i} \in \{0, \ldots, C-1 \}$, $\mathbbm{1}_{\{\cdot\}}(\cdot)$ is the indicator function that ensures the designated label flipping ratio of clean labels.

\subsection{Algorithm of BadLabel}
We craft a BadLabel dataset by solving Eq.~\eqref{eq:attack-objective} approximately. In particular, we need to find for which data the label is flipped and how its label should be flipped. 
However, this problem is optimization unfriendly because the label space is discrete.

Inspired by Zhao et al. (2017)~\cite{zhao2017efficient}, we introduce a \emph{flag} array $z \in \bR^{n \times C}$ that is initialized by the one-hot form of $n$ clean label $y_{i}$. The flag array will decide for which data the label is flipped and how its label should be flipped.

The magnitude of the element in the flag array $z$ should indicate the data's loss values to different classes. Therefore, we name $z(i,j)$ the $i$-th data's \textbf{affinity score} to the class $j$, where $i \in \{1,\ldots, n\} $ means the index of data and $j \in \{1,\ldots, C\}$ means the index of classes. 
For example, larger $z_T(i,j)$ means data $x_i$ with label $j$ has a smaller loss value over the models $\{f_1,\ldots, f_T\}$; smaller  $z_T(i,j)$ means data $x_i$ with label $j$ has a larger loss value over the models $\{f_1,\ldots, f_T\}$. 
To generate the BadLabel, we assign the $i$-th data a label $j$ with the lowest affinity score $z_T(i,j)$. 

To learn the flag array, we train a DNN for $T$ epochs. At every epoch $t$, we update the flag array $z$ as follows.
\begin{equation}
	\label{eq:solve-z}
	z_{t+1} = z_{t} - \alpha \nabla_{Y} \cL(Y, f_{t}(X)), 
\end{equation}
where $X$ is an $n\times d$ tensor (i.e., $[x_1,\ldots, x_n]^\top$), and $Y$ is an $n\times C$ array (i.e., one-hot version of hard labels), $t \in \{1, 2,\ldots, T \} $ is the iteration index, $f_t$ is a DNN at epoch $t$, and $\alpha$ is a small step size. In the following subsection~\ref{sec:explain_eq_badlabel}, we provide the rationality of Eq.~\eqref{eq:solve-z}.

Finally, the flag array $z_T$ at the last epoch $T$ will decide for which data the label is flipped and how its label should be flipped. 
Algorithm~\ref{alg:BadLabel} provides the details as follows. 

\begin{algorithm}[h!]
	\setstretch{1.05}
	\caption{Crafting the BadLabel}
	\label{alg:BadLabel}
	\begin{algorithmic}
		\STATE {\bfseries Input:}  A clean set $D=\{(x_{i}, y_{i})\}_{i=1}^{n}$, flipping ratio $\rho$, iteration $T$, step size $\alpha$.
		\STATE {\bfseries Output:}  A label-noise set $D'=\{(x_{i}, y_{i}')\}_{i=1}^{n}$.
		\STATE{//Stage I: Optimize the data's affinity score $z(i,j)$.}
		\STATE {Initialize flag array $z_{1} \in \bR^{n \times C} $} by $Y$ (i.e., one-hot version of $n$ clean label $y_i$).
		\FOR{epoch  $t = 1$, $\ldots$, $T$}
		\STATE {Iterate $D$ to optimize $f_t$ (see Eq.~\eqref{eq:dnns-objecetive}).}
		\STATE {Update $z_{t+1}$ by Eq.~\eqref{eq:solve-z}.}
		\STATE {Normalize $z_{t+1}$. // E.g. use softmax function}
		\ENDFOR
		\STATE {//Stage II: Obtain $z_{T}$ and flip $\rho$ ratio of labels.}
		\STATE{Re-arrange $D$ in ascending order by $\{\min z_{T}(i,:) \}_{i=1}^{n}$.}
		\STATE{//Select the first $\rho$ percentage of data.}
		\FOR{epoch  $i = 1$, $\ldots$, $\lfloor{\rho \times n}\rfloor $}
		\STATE{//Flip its label to the class with the lowest affinity score.}
		\STATE{$y_i' = \argmin z_{T}(i,:)$ }
		\ENDFOR
	\end{algorithmic}
\end{algorithm}

\subsection{Mathematical Analysis of BadLabel}
\label{sec:explain_eq_badlabel}
We mathematically explain Eq.~\eqref{eq:solve-z} used by Algorithm~\ref{alg:BadLabel}.
To maximize the loss in Eq.~\eqref{eq:attack-objective}, we aim to find a noisy label $Y'$ ($n \times C$ array and one-hot form of noisy labels) that maximizes the loss values over all the functions. 

Given a model $f_t$ at epoch $t$, we can use one-step optimization to approximately find $Y'_{t+1}$ that makes the loss value largest in model $f_t$, i.e.,
\begin{equation}
	\label{eq:optim_Y}
	Y'_{t+1} = Y'_{t} + \alpha \nabla_{Y} \ell(Y, f_{t}(X)),
\end{equation}
where $Y'_{t}$ is the last-epoch noisy label that approximately makes $\ell(Y'_{t}, f_{t-1}(X))$ largest; $Y$ is the true label. 

Then, we expand Eq.~\eqref{eq:optim_Y} over $T$ models of $T$ training epochs that approximately represent all models in the hypothesis space $\cF$, i.e., 
\begin{equation}
	\label{eq:optim_Y_expand}
	Y'_{T} = Y'_{0} + \alpha \sum\limits_{t=0}^{T} \nabla_{Y} \ell(Y, f_{t}(X)).
\end{equation}

To make $Y'_{T}$ correspond to an even larger loss value, we consider initializing $Y'_{0}$ in Eq.~\eqref{eq:optim_Y_expand} to be $-Y$ that negates the true label, which could be a reasonable starting point to maximize the loss:
\begin{equation}
	Y'_{T} = -Y + \alpha \sum\limits_{t=0}^{T} \nabla_{Y} \ell(Y, f_{t}(X)).
\end{equation}
Then, we let $z_{T} = -Y'_{T}$ and then derive the following equation:
\begin{equation}
	\label{eq:z_derive}
	z_{T} = Y - \alpha \sum\limits_{t=0}^{T} \nabla_{Y} \ell(Y, f_{t}(X)), 
\end{equation}
where Eq.~\eqref{eq:z_derive} corresponds to the optimization of the flag array $z$ in Eq.~\eqref{eq:solve-z}.

Finally, to meet the constraint on $\rho$, we leverage the flag array $z_{T}$ to select the $(100\times\rho)\%$ of data with the lowest affinity scores to flip their labels.

\subsection{Visualization of BadLabel}
In the top row of Figure~\ref{fig:prob_matrix_and_loss_dist}, we built a toy example to visualize and compare different types of label noise. We crafted a 3-class classification problem. We used points/squares/triangles to represent samples and different colors (green, red, blue) to represent different annotations on samples. Note that the dashed lines represent the true class boundaries that LNL methods aim to learn.

In Figure~\ref{fig:prob_matrix_and_loss_dist}(a), all samples are correctly labeled, and the learning algorithm can easily learn the true decision boundary and make the correct predictions. 
In Figure~\ref{fig:prob_matrix_and_loss_dist}(b), the labels are corrupted by symmetric noise that is uniformly distributed inside each class. 
In each class, the noisy labels are scattered and sparse, which hardly causes a significant impact on learning. Therefore, the large portion of correct labels will gradually guide the classifier to learn the true class boundaries. 
Figure~\ref{fig:prob_matrix_and_loss_dist}(c), we show the instance-dependent noise, where samples near the decision boundary are most likely to be wrongly annotated~\cite{zhang2020learning}. This type of noise may shift the classifier's decision boundary but will not ruin the learning completely. 

On the contrary, as shown in Figure~\ref{fig:prob_matrix_and_loss_dist}(d), BadLabel tends to flip the labels of samples that are far away from the class boundary, and the noisy labels are clustered together, which can easily mislead the learning algorithm to learn complex but wrong decision boundaries or even lead to the unstable training. As a result, BadLabel is a more challenging type of label noise compared to the existing types of label noise, which calls for robust LNL algorithms. 
In the next section, we provide a very first example of a robust LNL algorithm that can handle both the BadLabel and other existing types of label noise.

\section{Robust Label-noise Learning Algorithm}
\label{Sec:method_robust_dividemix}
In this section, we propose a robust LNL algorithm to handle the BadLabel dataset. 
We perturb labels and model the resulting loss values to select a labeled set $\cX$, which consists mostly of clean samples, and treat the rest as the unlabeled set $\cU$. 
Then, we use a semi-supervised learning (SSL) algorithm to train DNNs based on both $\cX$ and $\cU$. 
By doing so, we are able to mitigate the negative impact of BadLabel on the model's generalization performance.

\subsection{Key Observation under BadLabel dataset}
\begin{figure}[h!]
	\centering
	\subfigure[Before label perturbation at Epoch $\#5$]{
		\centering
		\includegraphics[scale=0.145]{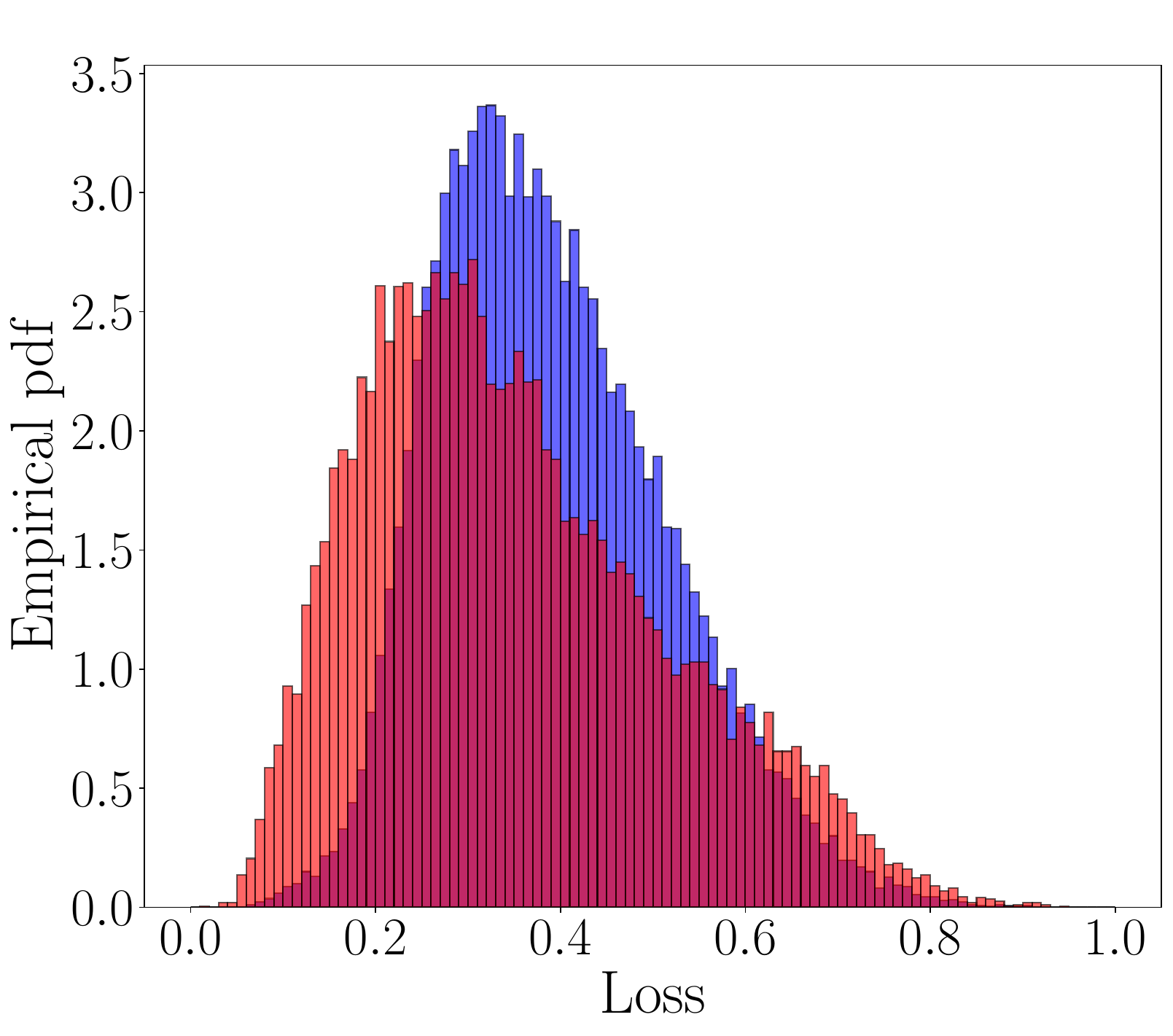}
		\label{fig:loss_dist_before_perturb}
	}	
	\hspace{2mm}
	\subfigure[After label perturbation at Epoch $\#5$]{
		\centering
		\includegraphics[scale=0.145]{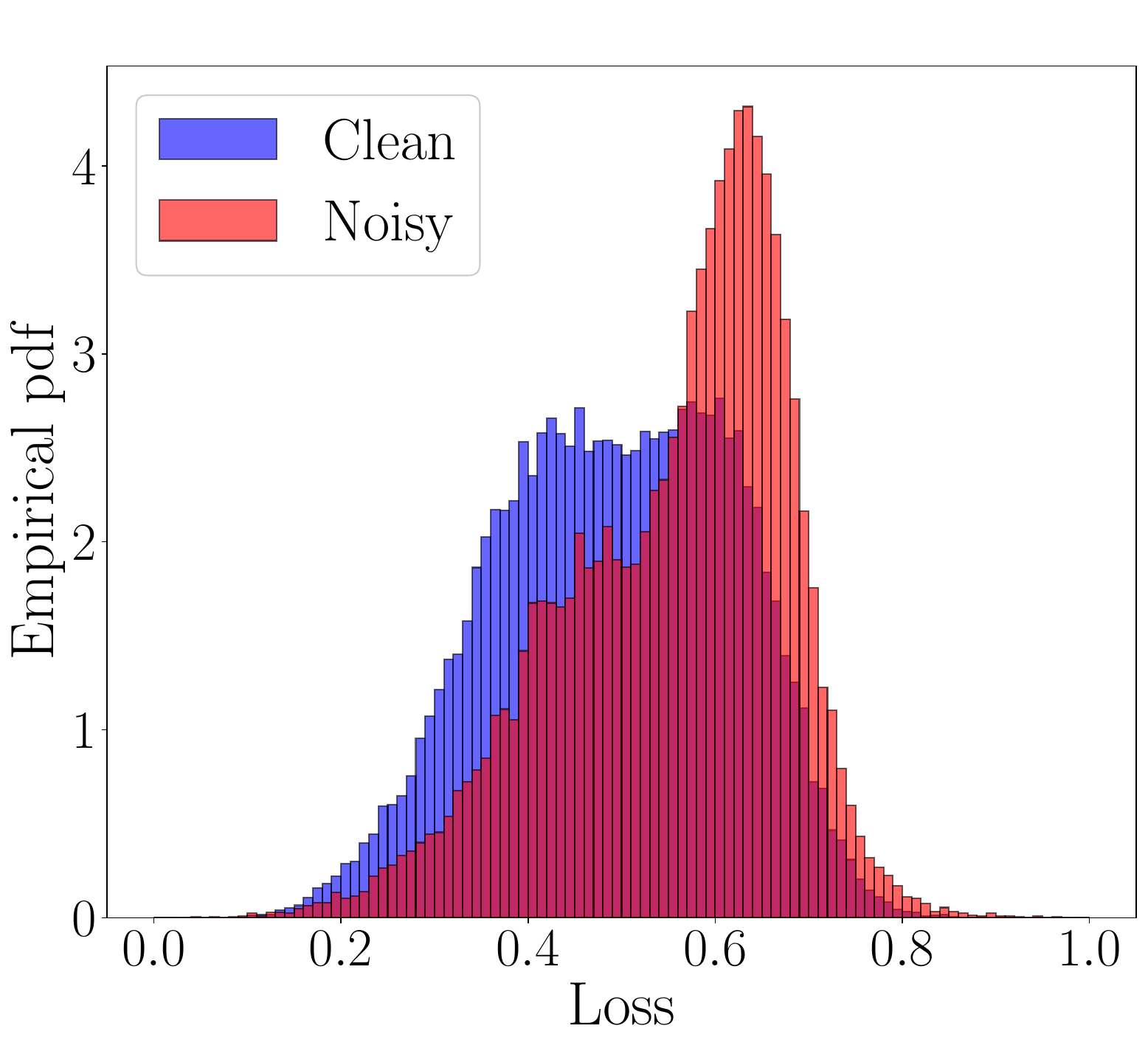}
		\label{fig:loss_dist_after_perturb}
	}
	\caption{On CIFAR-10 with $40\%$ BadLabel, we visualized the loss distribution of labels before and after label perturbations. After a few epoch warm-up training, (a) before adversarial perturbation of labels, noisy labels tend to have lower loss values; (b) after adversarial perturbation of labels, the noisy labels have larger loss values. In BadLabel, compared with clean labels, the noisy labels are more sensitive to adversarial perturbations. Note that we used hard labels to calculate the loss values despite the label perturbations.}
	\label{fig:loss_dist_perturbation}
\end{figure}

\noindent\textbf{DNNs fit noisy labels before clean labels in BadLabel.}
Previous studies~\cite{arpit2017closer,zhang2017understanding,chen2019understanding} have shown that DNNs tend to learn clean samples before noisy ones, resulting in clean samples having lower loss values than noisy samples, as shown in Figure~\ref{fig:prob_matrix_and_loss_dist}(b) and Figure~\ref{fig:prob_matrix_and_loss_dist}(c). 
This phenomenon is referred to as early learning~\cite{liu2020early}.
However, this phenomenon does not hold in the BadLabel dataset.
As shown in Figures~\ref{fig:loss_dist_perturbation}(a) (at Epoch $\#5$)  and~\ref{fig:prob_matrix_and_loss_dist}(d) (Epoch $\#10$), the DNNs fit noisy labels first in BadLabel, and the loss values of clean and noisy labels gradually become nearly indistinguishable.
Consequently, the conventional loss-based LNL methods~\cite{han2018co,li2019dividemix} become ineffective under the BadLabel dataset. 
Therefore, we need to find a new way to select clean labels and correct noisy labels.

\noindent\textbf{Adversarial perturbation of labels.} We make a key observation that after a few epochs of warm-up training, noisy labels have lower loss values than clean labels (shown in Figure~\ref{fig:loss_dist_before_perturb}), but are more susceptible to adversarial perturbations that increase their loss values (shown in Figure~\ref{fig:loss_dist_after_perturb}).

To perturb all training labels $Y'$ in an adversarial manner, we use the single-step adversarial perturbation formulation as shown in Eq.~\eqref{eq:perturb_Y}.
\begin{equation}
	\label{eq:perturb_Y}
	\tilde{Y} = Y' + \lambda \nabla_{Y'} \ell(Y', f(X)),
\end{equation}
where ${Y}' \in \bR^{n \times C}$ is an one hot form of noisy labels, $\tilde{Y} \in \bR^{n \times C}$ is an adversarially perturbed variant of ${Y}'$, and $\lambda$ is the step size.

The observation holds that \textit{before label perturbation, the loss values of noisy labels are smaller compared to those of clean labels, but after label perturbation, the loss values of noisy labels become larger.} This observation is crucial for the proposed LNL algorithm under the BadLabel scenario.


\subsection{Preliminary Techniques}
\noindent\textbf{Bayesian Gaussian Mixture Model (BayesGMM)~\cite{RobertsHRP1998bayesian}.} 
The standard Gaussian Mixture Model (GMM) is a clustering method, which can give a probability that each data belongs to each cluster.
Compared with the standard GMM, BayesGMM can infer from the data the most appropriate number of clusters. 
When BayesGMM models the per-sample loss distributions, BayesGMM converges slower than the standard GMM if the number of specified clusters is larger than the number of actual clusters~\cite{BayeGMM_speed}. 
Therefore, we use the convergence speed of BayesGMM to judge whether per-sample loss distributions of noisy and clean data can be properly differentiated. 
If BayesGMM fits $\{\ell(f_{\theta}(x_i),y'_i)\}^{n}_{i=1}$, where $(x_i, y'_i) \in D'$ and  $D'$ is a label-noise dataset, and the convergence speed is low, then we can infer that noisy and clean labels are indistinguishable, and conversely, if the convergence speed is fast, then we can infer that noisy and clean labels are distinguishable.

If BayesGMM takes input noisy training set $D'$ and network parameter $\theta$ and converges to $\delta$ within $N_{\mathrm{iter}}$ iterations, the output of BayesGMM is specified as follows. 
\begin{equation}
	\label{eq:BayesGMM}
	\cW  = {\rm BayesGMM}^{\delta, N_{\mathrm{iter}}}(D', \theta),
\end{equation}
where we specify $\cW \in [0,1]^n$ as a vector of size $n$ whose element $w_i$ is the posterior probability of smaller mean (data $i$-th smaller loss value). Given $N_{\mathrm{iter}}$ iterations, the convergence value $\delta$ judges the separability of clean and noisy labels.
%


\noindent\textbf{Confidence Penalty (CP)~\cite{pereyra2017regularizing}.} 
Li et al. (2019) proposed DivideMix~\cite{li2019dividemix}, which uses CP to improve the effectiveness of GMM in modeling clean and noisy labels. 
They achieved this by making the per-sample loss distributions more distinguishable. 
During training, a negative entropy term ($-\cH$) is added to the loss function Eq.~\eqref{eq:dnns-objecetive} to penalize overconfident predictions and increase the per-sample losses, where $\cH = - \sum\limits f_\theta(x){\log}(f_\theta(x))$. 
This prevents the per-sample losses from forming a cluster around zero, making it easier for GMM to separate the loss values.
%
%
%

\noindent\textbf{MixMatch~\cite{berthelot2019mixmatch}.} 
MixMatch is an effective SSL algorithm. 
MixMatch can effectively utilize unlabeled data by introducing the powerful data augmentation technique MixUp~\cite{mixup2018} and encouraging the network to make consistent and high-confident predictions on unlabeled data through consistency regularization and entropy minimization.
In DivideMix~\cite{li2019dividemix}, there are a pair of networks $k = 0$ or $1$. Given network $k$, GMM can separate the noisy dataset into (mostly) clean $\cX_k$ set and unlabeled set $\cU_k$. 
Then, MixMatch uses $\cX_k$ and $\cU_k$ as the training data to feed $k$'s peer network $(k-1)$. 
The per-epoch cross-training of MixMatch can be specified as 
\begin{equation}
	\label{eq:MixMatch}
	\theta_{1-k}^{e+1} = {\rm MixMatch}(\cX_{k}, \cU_{k}, \theta_{1-k}^{e}),
\end{equation}
where $e \in \{1,2,\ldots,E\}$ is the index of the total $E$ training epochs, and $k \in \{0, 1\}$ is the index of the pair networks $\theta_{k}$.


\begin{algorithm*}[h!]
	\setstretch{1.15}
	\caption{Robust DivideMix}
	\label{alg:new-robust-dividemix}
	\begin{algorithmic}
	\STATE {\bfseries Input:} a pair of DNNs parameterized by  $\theta_{1}, \theta_{2}$, noisy set $D'= (X, Y')$, selection threshold $\tau_p$ and $\tau_c$, MixMatch epochs $E$
		\STATE {\bfseries Output:} A pair of optimized DNNs parameterized by $\theta_1^{E}, \theta_2^{E}$ for making predictions jointly
		\STATE{//Stage I: Initialization of the pair of DNNs $\theta_1, \theta_2$}
		\STATE {$\theta_{1}^{0}, \theta_{2}^{0}$ = WarmUp($D', \theta_{1}, \theta_{2}$) // Use $D'$ to conduct standard training for a few epochs}
		\STATE {$\tilde{D}=\{(x_{i}, \tilde{y}_{i})\}_{i=1}^{n} \leftarrow$ perturb $y'_{i}$ by Eq.~\eqref{eq:perturb_Y}}
		\STATE {$\cW_{1}^p$ = BayesGMM($\tilde{D}, \theta_{2}^{0}$), $\cW_{2}^p$ = BayesGMM($\tilde{D}, \theta_{1}^{0}$)}
		\FOR {$k = 1,2$}
		\STATE{$\cX_k = \{(x_{i},y'_{i}) | w_i^p \geq \tau_p, \forall(x_{i}, y'_{i}, w_i^p) \in (X, Y', \cW_{k}^p)\}$ //  (mostly) clean labeled set}
		\STATE{$\cU_k = \{x_{i} |(x_i, y_i') \in D' \land (x_i, y_i') \not \in \cX_k \}$ // unlabeled set}
		\ENDFOR
		\STATE{Obtain $\theta_{1}^{1}$ and $\theta_{2}^{1}$ by Eq.~\eqref{eq:MixMatch} for a single epoch (i.e., $e=0 \to e=1$) // Obtain good initialization of the pair DNNs}
		\STATE{//Stage II: Pair-wise training of $\theta_1, \theta_2$ using BayesGMM and MixMatch for $E$ epochs}
		\STATE{$\cW_{1}^c = \cW_{2}^p, \cW_{2}^c = \cW_{1}^p$}
		\FOR {epoch $e = 1,\ldots,E$}
		\IF{BayesGMM($D', \theta_{2}^{e}$) is converged}
		\STATE{$\cW_{1}^c$ = BayesGMM($D', \theta_{2}^{e}$)}
		\ENDIF
		\IF{BayesGMM($D', \theta_{1}^{e}$) is converged}
		\STATE{$\cW_{2}^c$ = BayesGMM($D', \theta_{1}^{e}$)}
		\ENDIF
		\FOR {$k = 1,2$}
		\STATE{$\cX_k = \{(x_{i},y'_{i}) | w_i^c \geq \tau_c, \forall(x_{i}, y'_{i}, w_i^c) \in (X, Y', \cW_{k}^c)\}$ // (mostly) clean labeled set }
		\STATE{$\cU_k = \{x_{i} |(x_i, y_i') \in D' \land (x_i, y_i') \not \in \cX_k \}$ // unlabeled set}
		\ENDFOR
		\STATE{Obtain $\theta_{1}^{e+1}$ and $\theta_{2}^{e+1}$ by Eq.~\eqref{eq:MixMatch} (i.e., $e \to e+1$).}
		\ENDFOR
		\STATE{//Stage III: Predictions using the pair DNNs ($\theta^{E}_1$, $\theta^{E}_2$)}
		\STATE {$y = \argmax \big(f_{\theta^{E}_1} (x) + f_{\theta^{E}_2} (x) \big) $}
	\end{algorithmic}
\end{algorithm*} 

\subsection{Algorithm of Robust DivideMix}
Here, we provide a robust LNL method called Robust DivideMix (see Algorithm~\ref{alg:new-robust-dividemix}). 
Our method builds upon the preliminary techniques and can handle various types of label noise, including BadLabel noise. Robust DivideMix consists of three stages.

First, we properly initialize the pair networks. 
Specifically, we conduct warm-up training for a few epochs, in which we use CP to enhance the distinguishability of loss distributions.
Then, we leverage the above observation presented in Figure~\ref{fig:loss_dist_perturbation} to select a small set of (mostly) clean labeled data $\cX$ and treat the rest of data as unlabeled $\cU$.
We initialize the pair networks to $\theta^1_1$ and $\theta^1_2$ via one epoch of MixMatch cross-training Eq.~\eqref{eq:MixMatch}. 

%

Second, we leverage MixMatch to cross-update parameters of the pair DNNs for $E$ epochs. 
Unlike DivideMix~\cite{li2019dividemix} using GMM, our Robust DivideMix employs BayesGMM to model per-sample loss distributions.
When the loss distribution of BadLabel is unimodal rather than bimodal, the convergence speed of BayesGMM is slow when the number of components is preset to two.
The convergence speed enables us to determine whether clean and noisy labels can be effectively differentiated during the selection of clean labels.
After dividing the training set into labeled and unlabeled sets, we use MixMatch to cross-update the parameters of the pair DNNs.

Third, we leverage the two networks to make a joint prediction.

\noindent\textit{Remark}
In the first stage, we conduct warm-up training and partition the training data using label perturbation to obtain a high-quality labeled set that mostly contains clean labels. At this point, the DNN has not completely fit BadLabel, and the perturbation can effectively make the loss distributions distinguishable. However, in the second stage, we no longer use label perturbation because it can be difficult to control and can significantly harm training when noisy labels are accidentally selected. 
Therefore, we adopt BayesGMM to select clean labels and prevent noisy labels from being included in the labeled set.

\section{Experiments}
In this section, we present the results of our extensive experiments. 
We start by evaluating the impact of BadLabel on state-of-the-art LNL algorithms and demonstrate that it can significantly degrade their performance in Section~\ref{Sec:exp-BadLabel}

\begin{table*}[tp]
	\scriptsize
	\centering
	\renewcommand\arraystretch{1.1}
\caption{Test accuracy ($\%$) on CIFAR-10 with different types of label noise (symmetric, asymmetric, instance-dependent, and our proposed BadLabel) and noise levels (ranging from 20$\%$ to $80\%$). The most robust evaluations for each LNL method are highlighted in bold.}
	\label{exp:cifar10-BadLabel}
	\setlength{\tabcolsep}{1.4mm}{
		\begin{tabular}{lc|cccc|cc|cccc|cccc}
			\toprule
			\multirow{3}*{Method}&\multirow{3}*{}&\multicolumn{14}{c}{Noise Type / Noise Ratio} \\
			\cmidrule{3-16}  &&\multicolumn{4}{c|}{Sym.}&\multicolumn{2}{c|}{Asym.}&\multicolumn{4}{c|}{IDN}&\multicolumn{4}{c}{BadLabel} \\	
			&&20\%&40\%&60\%&80\%&20\%&40\%&20\%&40\%&60\%&80\%&20\%&40\%&60\%&80\% \\
			\midrule
			\multirow{2}*{Standard Training}
			&Best&85.21&79.90&69.79&43.00&88.02&85.22&85.42&78.93&68.97&55.34&\textbf{76.76}$\pm$1.08&\textbf{58.79}$\pm$1.49&\textbf{39.64}$\pm$1.13&\textbf{17.80}$\pm$0.91\\
			&Last&82.55&64.79&41.43&17.20&87.28&77.04&85.23&74.06&52.22&28.04&\textbf{75.31}$\pm$0.24&\textbf{55.72}$\pm$0.17&\textbf{35.66}$\pm$0.23&\textbf{13.44}$\pm$0.26\\
			\midrule
			Co-teaching&Best&89.19&84.80&58.25&21.76&90.65&63.11&85.72&73.42&45.84&33.43&\textbf{80.41}$\pm$0.78&\textbf{56.81}$\pm$3.86&\textbf{14.42}$\pm$1.22&\textbf{10.51}$\pm$0.71\\
			Han et al. (2018)~\cite{han2018co}&Last&89.03&84.65&57.95&21.06&90.52&56.33&85.48&72.97&45.53&25.27&\textbf{79.48}$\pm$0.75&\textbf{55.54}$\pm$3.74&\textbf{12.99}$\pm$1.09&\textbf{4.24}$\pm$2.44\\
			\midrule
			T-Revision&Best&89.79&86.83&78.14&64.54&91.23&89.60&85.74&78.45&69.31&56.26&\textbf{76.99}$\pm$1.38&\textbf{57.21}$\pm$1.64&\textbf{36.01}$\pm$1.10&\textbf{14.93}$\pm$0.50\\
			Xia et al. (2019)~\cite{xia2019anchor}&Last&89.59&86.57&76.85&60.54&91.09&89.40&85.43&69.18&58.15&33.15&\textbf{75.71}$\pm$1.68&\textbf{55.02}$\pm$1.34&\textbf{33.99}$\pm$0.29&\textbf{13.16}$\pm$0.68\\
			\midrule
			RoG&Best&-&-&-&-&-&-&-&-&-&-&-&-&-&-\\
			Lee et al. (2019)~\cite{lee2019robust}&Last&87.48&74.81&52.42&16.02&89.61&81.63&\textbf{85.34}&76.68&63.79&37.11&85.88$\pm$0.32&\textbf{64.20}$\pm$0.91&\textbf{35.89}$\pm$1.34&\textbf{8.64}$\pm$0.76\\
			\midrule
			DivideMix&Best&96.21&95.08&94.80&81.95&94.82&94.20&91.97&85.84&81.59&59.06&\textbf{84.81}$\pm$0.78&\textbf{58.44}$\pm$1.45&\textbf{28.38}$\pm$0.56&\textbf{6.87}$\pm$0.59\\
			Li et al. (2019)~\cite{li2019dividemix}&Last&96.04&94.74&94.56&81.58&94.46&93.50&90.77&82.94&81.19&47.81&\textbf{82.13}$\pm$0.78&\textbf{57.65}$\pm$1.96&\textbf{16.21}$\pm$1.24&\textbf{6.12}$\pm$0.45\\
			\midrule
			AdaCorr&Best&90.66&87.17&80.97&35.97&92.35&88.60&85.88&79.54&69.36&55.86&\textbf{76.97}$\pm$0.83&\textbf{57.17}$\pm$0.71&\textbf{37.14}$\pm$0.38&\textbf{14.72}$\pm$0.86\\
			Zheng et al. (2020)~\cite{zheng2020error}&Last&90.46&86.78&80.66&35.67&92.17&88.34&85.70&79.05&59.13&30.48&\textbf{74.71}$\pm$0.26&\textbf{54.92}$\pm$0.22&\textbf{34.71}$\pm$0.22&\textbf{11.94}$\pm$0.12\\
			\midrule
			Peer Loss&Best&90.87&87.13&79.03&61.91&91.47&87.50&86.46&81.07&69.87&55.51&\textbf{75.28}$\pm$1.43&\textbf{55.75}$\pm$1.39&\textbf{36.17}$\pm$0.23&\textbf{15.87}$\pm$0.30\\
			Liu et al. (2020)~\cite{liu2020peer}&Last&90.65&86.85&78.83&61.43&91.11&81.24&85.72&74.43&54.57&33.76&\textbf{74.00}$\pm$1.43&\textbf{53.73}$\pm$1.25&\textbf{34.37}$\pm$0.68&\textbf{14.71}$\pm$0.22\\
			\midrule ELR&Best&92.85&91.30&87.99&54.67&92.42&89.40&87.62&82.08&73.23&57.26&\textbf{85.73}$\pm$0.15&\textbf{62.58}$\pm$1.33&\textbf{35.24}$\pm$1.12&\textbf{11.71}$\pm$0.70\\
			Liu et al. (2020)~\cite{liu2020early}&Last&89.37&87.78&85.69&46.71&92.31&89.11&85.31&78.05&68.12&48.99&\textbf{81.88}$\pm$0.25&\textbf{56.45}$\pm$0.31&\textbf{30.45}$\pm$0.30&\textbf{8.67}$\pm$0.79\\
			\midrule
			Negative LS&Best&87.42&84.40&75.22&43.62&88.34&85.03&89.82&83.66&75.76&64.21&\textbf{78.77}$\pm$0.66&\textbf{57.68}$\pm$0.89&\textbf{36.57}$\pm$0.88&\textbf{16.46}$\pm$0.82\\
			Wei et al. (2021)~\cite{wei2021understanding}&Last&87.30&84.21&75.07&43.50&\textbf{65.23}&\textbf{47.22}&81.87&82.10&70.95&45.62&73.99$\pm$0.90&52.45$\pm$1.03&\textbf{26.66}$\pm$0.81&\textbf{3.21}$\pm$0.44\\
			\midrule
			PGDF&Best&96.63&96.12&95.05&80.69&96.05&89.87&91.81&85.75&76.84&59.60&\textbf{82.72}$\pm$0.47&\textbf{61.50}$\pm$1.87&\textbf{34.46}$\pm$1.44&\textbf{6.37}$\pm$0.34\\
			Chen et al. (2021)~\cite{chen2021sample}&Last&96.40&95.95&94.75&79.76&95.74&88.45&91.30&84.31&69.54&34.81&\textbf{79.95}$\pm$0.36&\textbf{56.26}$\pm$1.03&\textbf{30.14}$\pm$0.85&\textbf{4.56}$\pm$0.45\\
			\midrule
			ProMix&Best&97.40&96.98&90.80&61.15&97.04&96.09&94.72&91.32&76.22&54.01&\textbf{94.95}$\pm$1.43&\textbf{48.36}$\pm$1.72&\textbf{24.87}$\pm$1.47&\textbf{9.51}$\pm$1.51\\
			Wang et al. (2022)~\cite{wang2022promix}&Last&97.30&96.91&90.72&52.25&96.94&96.03&94.63&91.01&75.12&45.80&\textbf{94.59}$\pm$1.64&\textbf{44.08}$\pm$0.49&\textbf{21.33}$\pm$0.46&\textbf{7.93}$\pm$1.34\\
			\midrule
			SOP&Best&96.17&95.64&94.83&89.94&95.96&93.60&90.32&83.26&71.54&57.14&\textbf{84.96}$\pm$0.35&\textbf{66.25}$\pm$1.35&\textbf{42.59}$\pm$1.25&\textbf{12.70}$\pm$0.89\\
			Liu et al. (2022)~\cite{liu22sop}&Last&96.12&95.46&94.71&89.78&95.86&93.30&90.13&82.91&63.14&29.86&\textbf{82.64}$\pm$0.27&\textbf{61.89}$\pm$0.25&\textbf{36.51}$\pm$0.26&\textbf{8.63}$\pm$0.17\\
			\midrule
			Robust DivideMix&Best&95.45&94.84&94.25&61.59&91.77&86.88&\textbf{90.44}&89.71&78.12&60.64&92.07$\pm$1.06&\textbf{86.70}$\pm$3.83&\textbf{76.47}$\pm$3.89&\textbf{27.41}$\pm$3.25\\
			Ours&Last&95.28&94.71&94.11&60.98&90.62&\textbf{84.02}&\textbf{87.30}&89.16&\textbf{72.33}&50.38&91.76$\pm$1.27&85.96$\pm$4.33&73.29$\pm$3.81&\textbf{25.20}$\pm$2.72\\
			\bottomrule 
	\end{tabular}}
 \vspace{-2mm}
\end{table*}

In Section~\ref{Sec:exp-robust-dividemix}, we show that our proposed Robust DivideMix method can handle various types of label noise, including BadLabel. 
We compare our method to existing LNL methods on different benchmark datasets and show that it outperforms or matches them in terms of accuracy. All experiments were conducted using a single NVIDIA TESLA V100 GPU.
\footnote{To save space, we report only the mean accuracy and standard deviation of our proposed methods. For existing methods, we faithfully use the official codes and refer interested readers to the original papers for their standard deviations.}



\subsection{BadLabel}
\label{Sec:exp-BadLabel}
In this subsection, we evaluate BadLabel on 11 state-of-the-art LNL algorithms: Co-teaching~\cite{han2018co}, T-Revision~\cite{xia2019anchor}, RoG~\cite{lee2019robust}, DivideMix~\cite{li2019dividemix}, AdaCorr~\cite{zheng2020error}, Peer Loss~\cite{liu2020peer}, ELR~\cite{liu2020early}, Negative LS~\cite{wei2021understanding}, PGDF~\cite{chen2021sample}, ProMix~\cite{wang2022promix}, SOP~\cite{liu22sop}. As a baseline, we also report the evaluation results on Standard Training, which only uses the cross-entropy loss function for the vanilla training.
We conduct experiments on CIFAR-10, CIFAR-100~\cite{krizhevsky2009learning}, and MNIST~\cite{lecun-mnist} datasets. 

We compare BadLabel with three commonly used synthetic noise types: symmetric noise (Sym.), asymmetric noise (Asym.), and instance-dependent noise (IDN). For symmetric noise, we randomly flip the true label to other classes. For instance-dependent noise, we follow the approach proposed by Chen et al. (2021)~\cite{chen2021beyond} for generating the label noise.

We generate the BadLabel datasets of CIFAR-10/100 by utilizing the PreAct-ResNet18~\cite{he2016identity} backbone in Algorithm~\ref{alg:BadLabel}. The network is trained using the cross-entropy loss function and the SGD optimizer with a momentum of 0.9 and a weight decay of 0.0005. Iteration $T$ is set to 120. The initial learning rate is set at 0.1, which decreases by a factor of 10 at the 60th and 90th iterations respectively. We set the step size $\alpha$ to 0.1. 
To ensure consistency, for all LNL algorithms, we use the same network architecture and random seed. In this section, the backbone we adopt for LNL is PreAct-ResNet18, and we also report the evaluation results using DenseNet~\cite{huang2017densely} as the backbone in the Appendix~\ref{Sec:densenet-BadLabel}.

Tables~\ref{exp:cifar10-BadLabel} and \ref{exp:cifar100-BadLabel} show the evaluation results of various LNL algorithms on CIFAR-10 and CIFAR-100 with different noise types and ratios.
We report the best test accuracy across all epochs (Best) and the average test accuracy over the last 10 epochs (Last). For each experiment on BadLabel, we repeat it five times with different random seeds to obtain a standard deviation. As shown in Table~\ref{exp:cifar10-BadLabel}, BadLabel significantly degrades the performance of all LNL algorithms by a large margin. Although most methods can effectively deal with the conventional label noise of various noise ratios, it is challenging to deal with BadLabel dataset. Especially at high noise ratios (60\%, 80\%), BadLabel almost ruined the training. This shows that BadLabel is more challenging than conventional synthetic noise. In other words, BadLabel can robustly evaluate the existing LNL algorithms, which also calls for more robust LNL methods. 
We also report similar results on MNIST in Appendix~\ref{Sec:cifar100-mnist-BadLabel}.

\begin{table*}[t!]
	\scriptsize
	\centering
	\renewcommand\arraystretch{1.1}
	\caption{
 Test accuracy ($\%$) on CIFAR-100 with different types of label noise (symmetric, instance-dependent, and our proposed BadLabel) and noise levels (ranging from 20$\%$ to $80\%$). The most robust evaluations for each LNL method are highlighted in bold.}
	\label{exp:cifar100-BadLabel}
	\setlength{\tabcolsep}{1.8mm}{
		\begin{tabular}{lc|cccc|cccc|cccc}
			\toprule
			\multirow{3}*{Method}&\multirow{3}*{}&\multicolumn{12}{c}{Noise Type / Noise Ratio} \\
			\cmidrule{3-14}  &&\multicolumn{4}{c|}{Sym.}&\multicolumn{4}{c|}{IDN}&\multicolumn{4}{c}{BadLabel} \\	
			&&20\%&40\%&60\%&80\%&20\%&40\%&60\%&80\%&20\%&40\%&60\%&80\% \\
			\midrule
			\multirow{2}*{Standard Training}
			&Best&61.41&51.21&38.82&19.89&70.06&62.48&53.21&45.77&\textbf{56.75}$\pm$0.98&\textbf{35.42}$\pm$0.77&\textbf{17.70}$\pm$1.02&\textbf{6.03}$\pm$0.24\\
			&Last&61.17&46.27&27.01&9.27&69.94&62.32&52.55&40.45&\textbf{56.30}$\pm$0.13&\textbf{34.90}$\pm$0.17&\textbf{17.05}$\pm$0.28&\textbf{4.18}$\pm$0.16\\
			\midrule
			Co-teaching&Best&62.80&55.02&34.66&7.72&66.16&57.55&45.38&23.83&\textbf{54.30}$\pm$0.78&\textbf{26.02}$\pm$2.13&\textbf{3.97}$\pm$0.11&\textbf{0.99}$\pm$0.21\\
			Han et al. (2018)~\cite{han2018co}&Last&62.35&54.84&33.44&6.78&66.02&57.33&45.24&23.72&\textbf{53.97}$\pm$0.71&\textbf{25.74}$\pm$1.21&\textbf{3.67}$\pm$0.14&\textbf{0.00}$\pm$0.00\\
			\midrule
			T-Revision&Best&65.19&60.43&43.01&4.03&68.77&62.86&54.23&45.67&\textbf{57.86}$\pm$1.02&\textbf{40.60}$\pm$1.33&\textbf{13.06}$\pm$1.20&\textbf{1.92}$\pm$0.56\\
			Xia et al. (2019)~\cite{xia2019anchor}&Last&64.95&60.26&42.77&3.12&68.53&62.39&53.07&41.85&\textbf{57.26}$\pm$1.54&\textbf{38.40}$\pm$0.96&\textbf{12.65}$\pm$0.58&\textbf{1.43}$\pm$0.95\\
			\midrule
			RoG&Best&-&-&-&-&-&-&-&-&-&-&-&-\\
			Lee et al. (2019)~\cite{lee2019robust}&Last&66.68&60.79&53.08&22.73&\textbf{66.39}&60.80&56.00&48.62&70.55$\pm$0.55&\textbf{58.61}$\pm$0.65&\textbf{25.74}$\pm$0.28&\textbf{4.13}$\pm$0.41\\
			\midrule
			DivideMix&Best&77.36&75.02&72.25&57.56&72.79&67.82&61.08&51.50&\textbf{65.55}$\pm$0.65&\textbf{42.72}$\pm$0.44&\textbf{19.17}$\pm$1.28&\textbf{4.67}$\pm$0.87\\
			Li et al. (2019)~\cite{li2019dividemix}&Last&76.87&74.66&71.91&57.08&72.50&67.37&60.55&47.86&\textbf{64.96}$\pm$0.47&\textbf{40.92}$\pm$0.36&\textbf{13.04}$\pm$0.85&\textbf{1.10}$\pm$0.21\\
			\midrule
			AdaCorr&Best&66.31&59.78&47.22&24.15&68.89&62.63&54.91&45.22&\textbf{56.22}$\pm$0.82&\textbf{35.38}$\pm$1.27&\textbf{16.87}$\pm$1.36&\textbf{4.81}$\pm$0.22\\
			Zheng et al. (2020)~\cite{zheng2020error}&Last&66.03&59.48&47.04&23.90&68.72&62.45&54.68&41.95&\textbf{55.69}$\pm$0.44&\textbf{33.88}$\pm$0.88&\textbf{14.88}$\pm$0.52&\textbf{3.76}$\pm$1.24\\
			\midrule
			Peer Loss&Best&61.97&51.09&39.98&18.82&69.63&63.32&55.01&46.20&\textbf{55.58}$\pm$1.79&\textbf{37.11}$\pm$2.01&\textbf{19.53}$\pm$1.29&\textbf{6.42}$\pm$0.52\\
			Liu et al. (2020)~\cite{liu2020peer}&Last&60.64&43.64&26.23&7.65&69.38&62.70&53.90&42.14&\textbf{55.00}$\pm$1.41&\textbf{35.85}$\pm$1.48&\textbf{18.65}$\pm$0.22&\textbf{5.74}$\pm$0.76\\
			\midrule
			ELR&Best&72.25&68.75&60.01&26.89&70.27&66.04&60.59&52.81&\textbf{68.21}$\pm$0.62&\textbf{43.75}$\pm$0.21&\textbf{14.39}$\pm$0.35&\textbf{1.09}$\pm$0.18\\
			Liu et al. (2020)~\cite{liu2020early}&Last&72.13&68.60&59.78&23.95&70.13&65.87&60.41&52.57&\textbf{67.97}$\pm$0.17&\textbf{43.40}$\pm$0.22&\textbf{13.97}$\pm$0.38&\textbf{0.98}$\pm$0.11\\
			\midrule
			Negative LS&Best&63.65&57.17&44.18&21.31&69.20&62.67&54.49&46.96&\textbf{57.76}$\pm$0.56&\textbf{36.80}$\pm$0.21&\textbf{17.96}$\pm$0.31&\textbf{5.88}$\pm$0.11\\
			Wei et al. (2021)~\cite{wei2021understanding}&Last&63.54&56.98&43.98&21.19&63.38&55.72&42.87&24.69&\textbf{56.42}$\pm$0.71&\textbf{33.38}$\pm$0.22&\textbf{11.42}$\pm$0.38&\textbf{1.28}$\pm$0.14\\
			\midrule
			PGDF&Best&81.90&78.50&74.05&52.48&75.87&71.72&62.76&53.16&\textbf{69.44}$\pm$0.26&\textbf{46.39}$\pm$0.39&\textbf{19.05}$\pm$0.37&\textbf{5.08}$\pm$0.13\\
			Chen et al. (2021)~\cite{chen2021sample}&Last&81.37&78.21&73.64&52.11&74.90&71.32&62.06&51.68&\textbf{68.18}$\pm$0.16&\textbf{45.38}$\pm$0.15&\textbf{16.84}$\pm$0.24&\textbf{0.72}$\pm$0.25\\
			\midrule
			ProMix&Best&79.99&80.21&71.44&44.97&76.61&71.92&66.04&51.96&\textbf{69.80}$\pm$1.58&\textbf{37.73}$\pm$1.09&\textbf{15.92}$\pm$1.88&\textbf{4.62}$\pm$0.95\\
			Wang et al. (2022)~\cite{wang2022promix}&Last&79.77&79.95&71.25&44.64&76.44&71.66&65.94&51.77&\textbf{69.68}$\pm$0.99&\textbf{37.24}$\pm$0.84&\textbf{14.88}$\pm$1.02&\textbf{3.42}$\pm$0.22\\
			\midrule
			SOP&Best&77.35&75.20&72.39&63.13&72.52&63.84&56.79&50.20&\textbf{65.80}$\pm$0.68&\textbf{45.61}$\pm$0.34&\textbf{22.68}$\pm$0.27&\textbf{2.88}$\pm$0.11\\
			Liu et al. (2022)~\cite{liu22sop}&Last&77.11&74.89&72.10&62.87&72.11&63.15&53.35&40.77&\textbf{65.51}$\pm$0.12&\textbf{45.24}$\pm$0.26&\textbf{21.55}$\pm$0.18&\textbf{2.48}$\pm$0.16\\
			\midrule
			Robust DivideMix&Best&77.35&74.40&70.74&48.13&73.49&69.47&63.64&52.74&\textbf{65.29}$\pm$0.76&\textbf{46.64}$\pm$0.48&\textbf{41.80}$\pm$1.19&\textbf{21.48}$\pm$0.39\\
			Ours&Last&77.06&74.16&69.93&47.84&73.10&68.88&61.03&46.84&\textbf{64.49}$\pm$0.96&\textbf{45.26}$\pm$0.40&\textbf{35.91}$\pm$0.67&\textbf{16.91}$\pm$0.41\\
			\bottomrule 
	\end{tabular}}
 \vspace{-2mm}
\end{table*}

\subsection{Robust DivideMix}
\label{Sec:exp-robust-dividemix}

In this subsection, we robustly evaluate and compare various LNL algorithms, including our proposed Robust DivideMix, using the BadLabel datasets of CIFAR-10 and CIFAR-100. 
Additionally, we examine the generalization of Robust DivideMix and standard DivideMix under conventional synthetic noises such as symmetric and instance-dependent label noises. 
Furthermore, we also evaluate the generalization of Robust DivideMix on real-world noise datasets such as CIFAR-10N~\cite{wei2021learning} and Clothing1M~\cite{xiao2015learning}. For each experiment, we repeatedly run Robust DivideMix five times using different random seeds.


\begin{figure*}[t!]
	\centering
	\subfigure[20\% BadLabel]{
		\centering
		\includegraphics[scale=0.26]{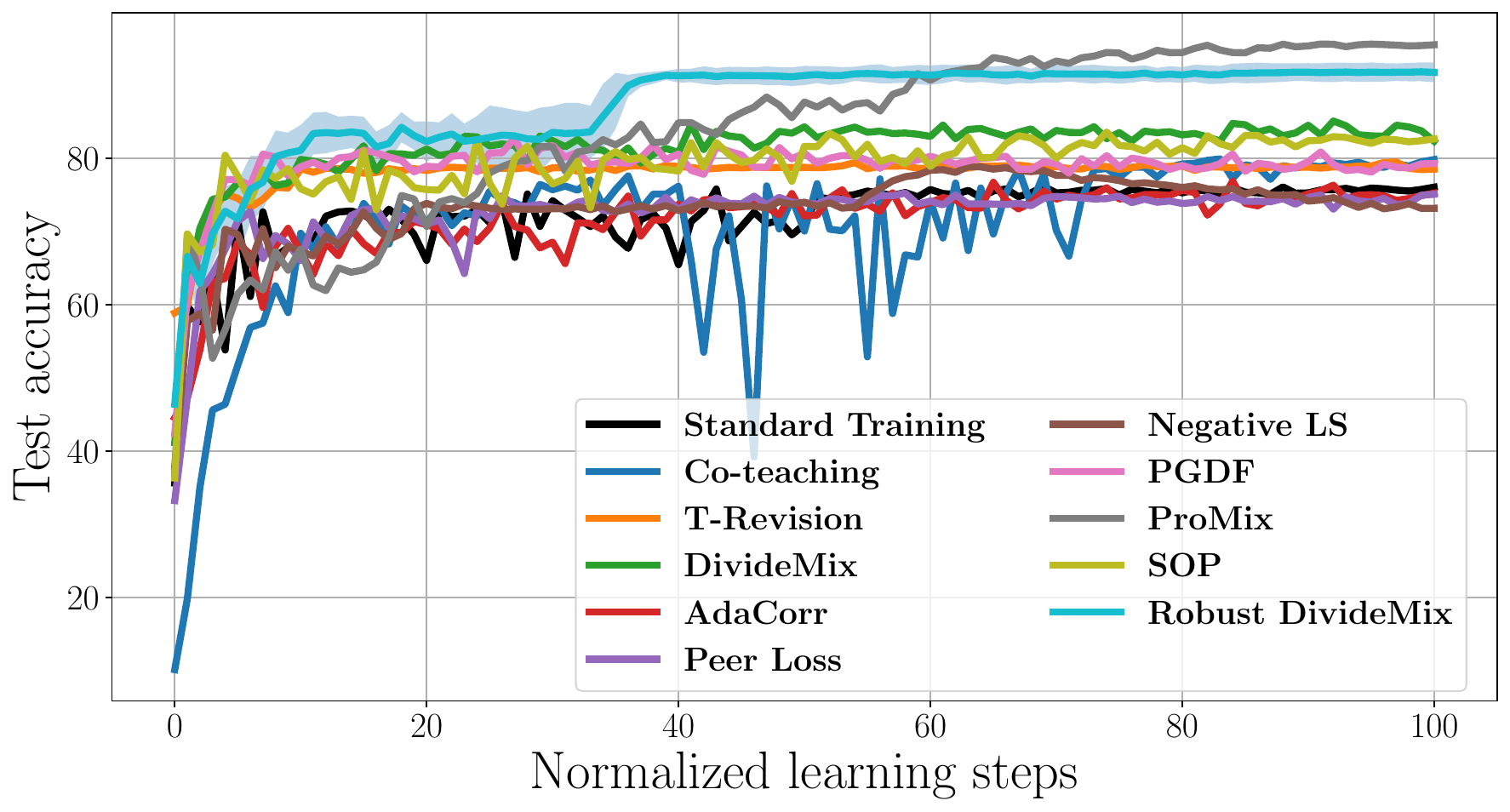}
		\label{fig:robdm_bad20}
	}	
	\hspace{5mm}
	\subfigure[40\% BadLabel]{
		\centering
		\includegraphics[scale=0.26]{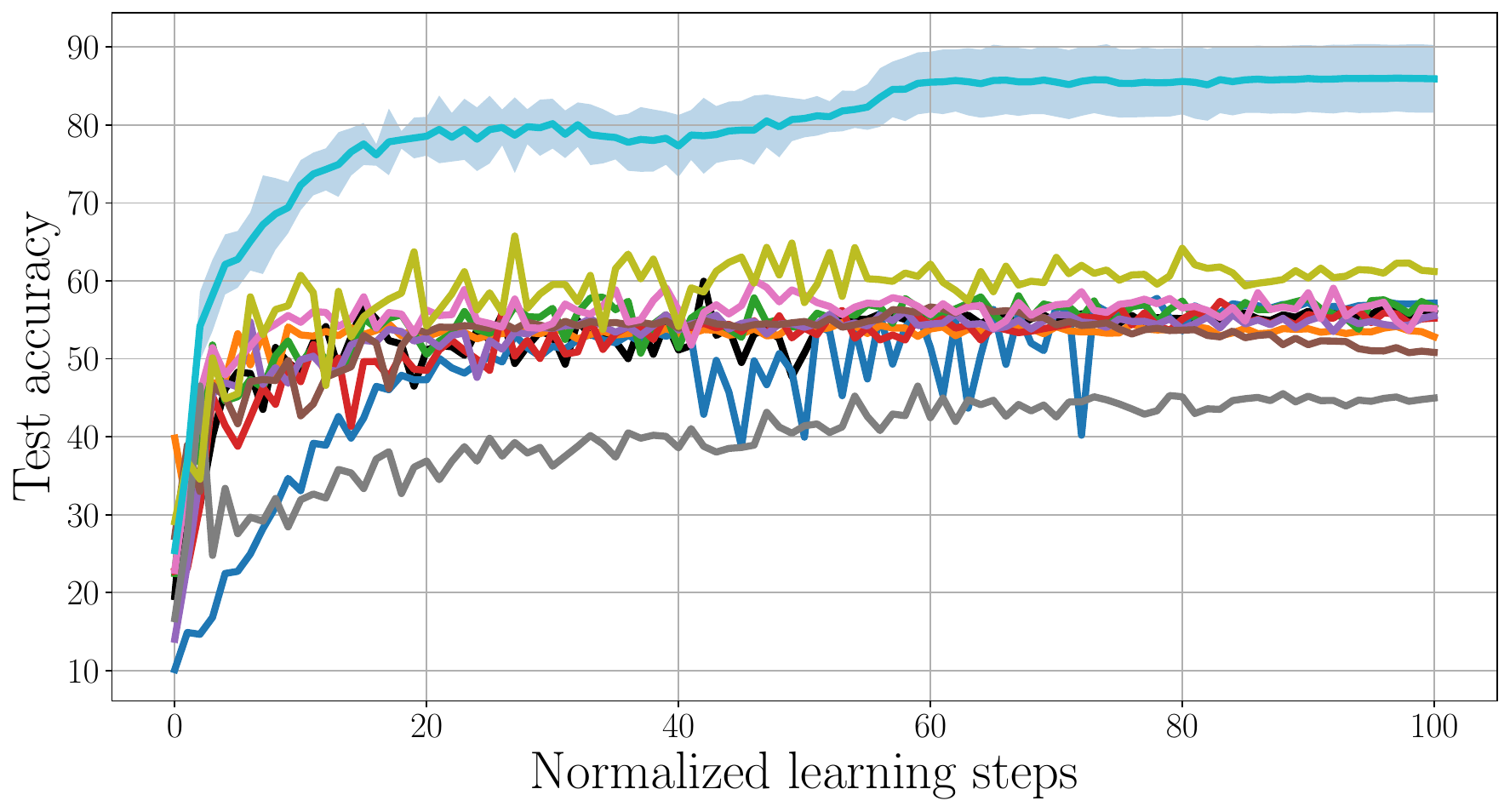}
		\label{fig:robdm_bad40}
	}
	\\
	\vspace{-2mm}
	\subfigure[60\% BadLabel]{
		\centering
		\includegraphics[scale=0.26]{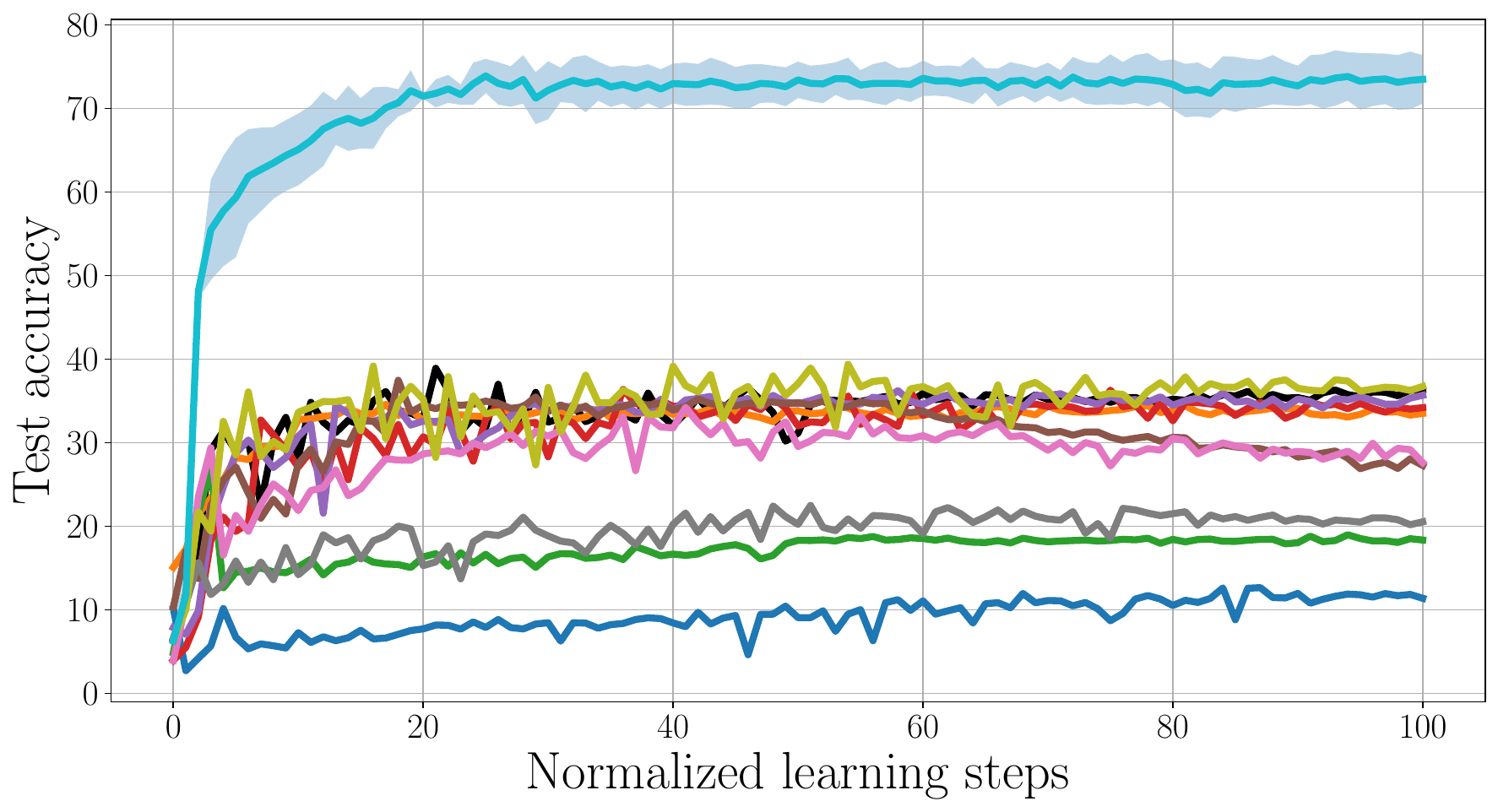}
		\label{fig:robdm_bad60}
	}
	\hspace{5mm}
	\subfigure[80\% BadLabel]{
		\centering
		\includegraphics[scale=0.26]{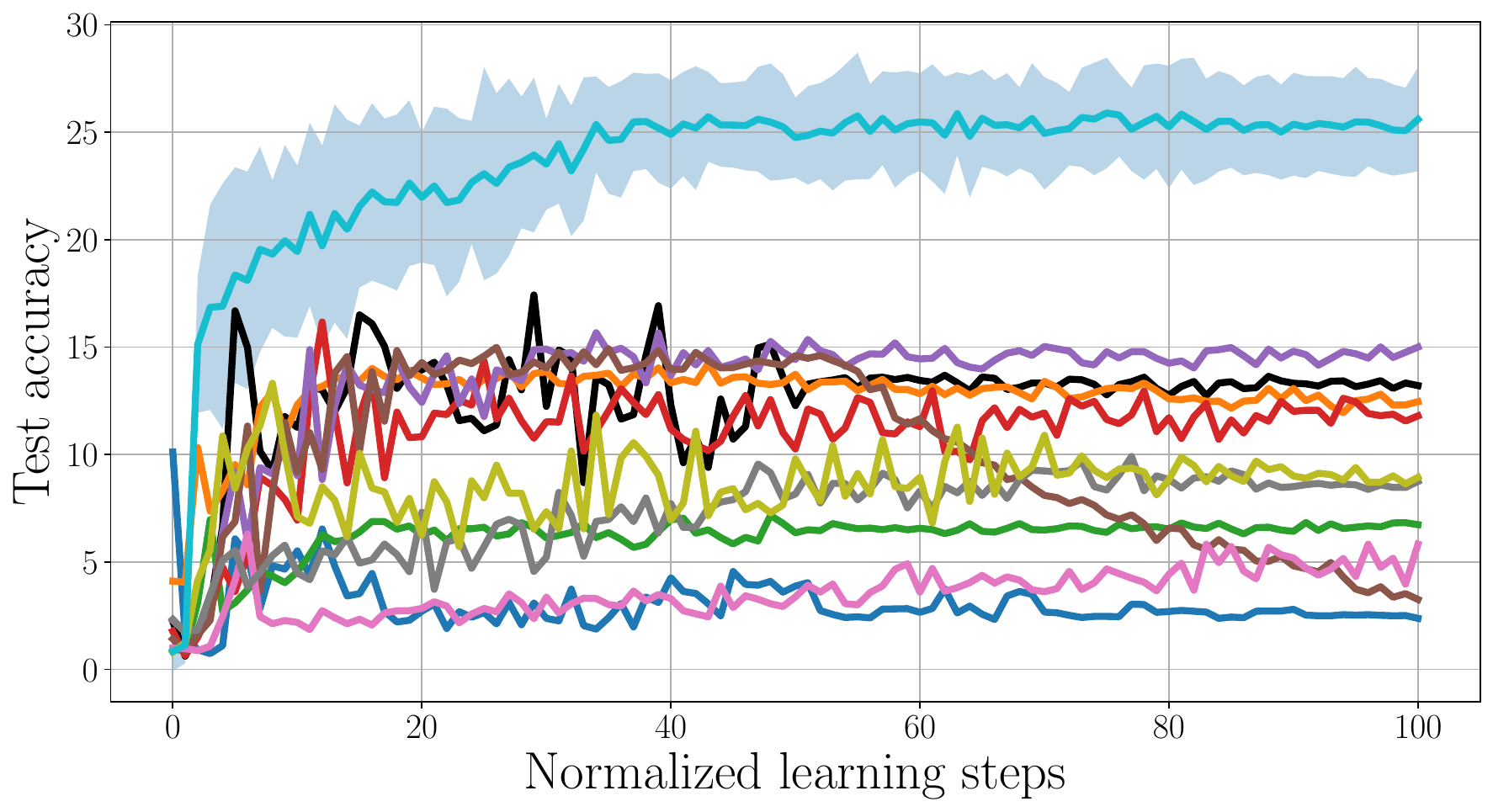}
		\label{fig:robdm_bad80}
	}
	\vspace{-2mm}
	\caption{Learning curves of several LNL algorithms on CIFAR-10 under varying BadLabel noise ratios. The shaded area represents the error bar corresponding to the standard deviation of Robust DivideMix.
    Note that, to facilitate a fair comparison of the learning curves, we normalized the learning steps by using uniform sampling, taking into account that different LNL algorithms have different optimal learning schedules.}
    \vspace{-2mm}
	\label{fig:robdm_learning_curve}
\end{figure*}

\begin{figure*}[h!]
	\centering
	\subfigure[20\% BadLabel]{
		\centering
		\includegraphics[scale=0.26]{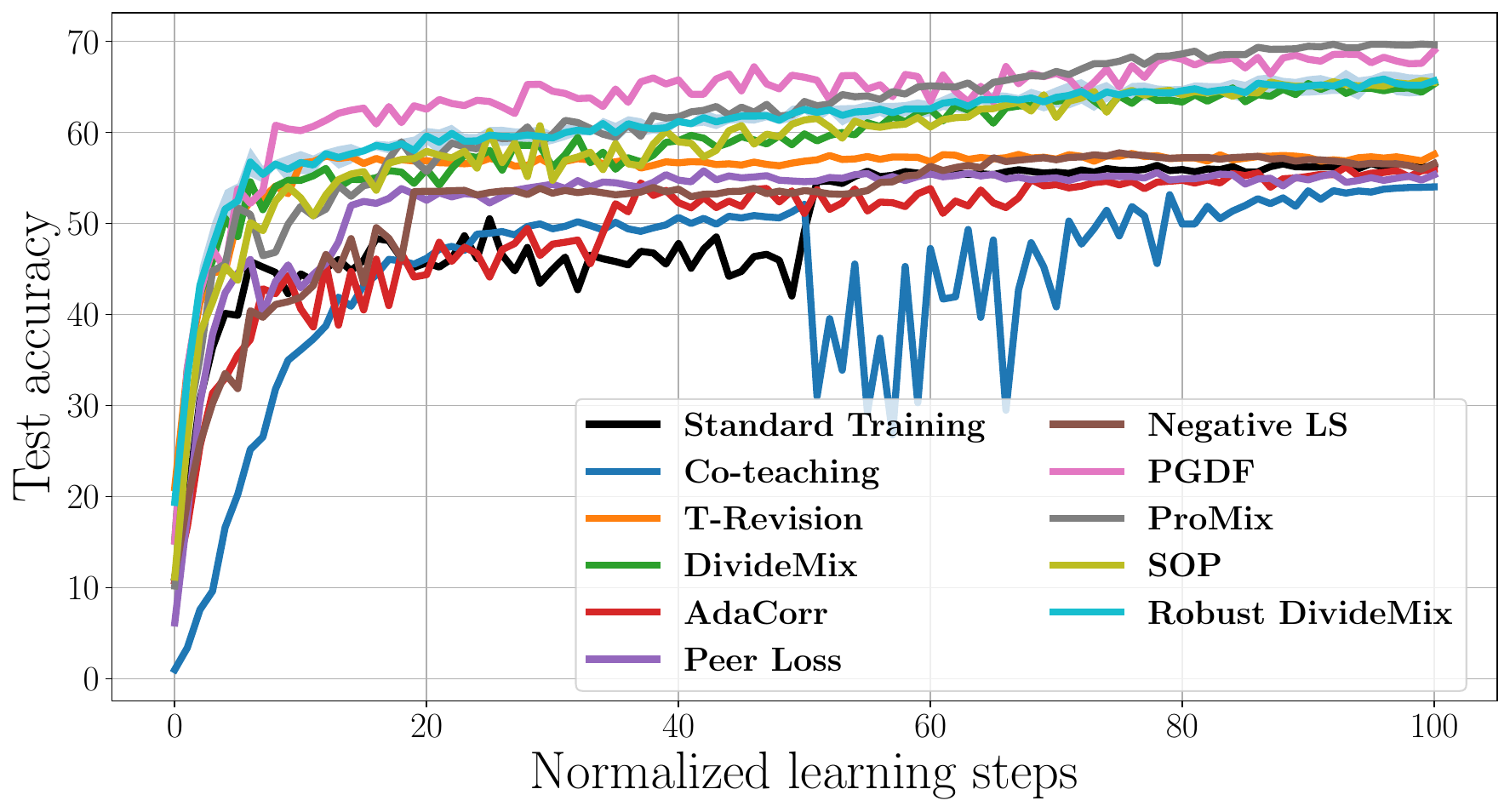}
	}	
	\hspace{5mm}
	\subfigure[40\% BadLabel]{
		\centering
		\includegraphics[scale=0.26]{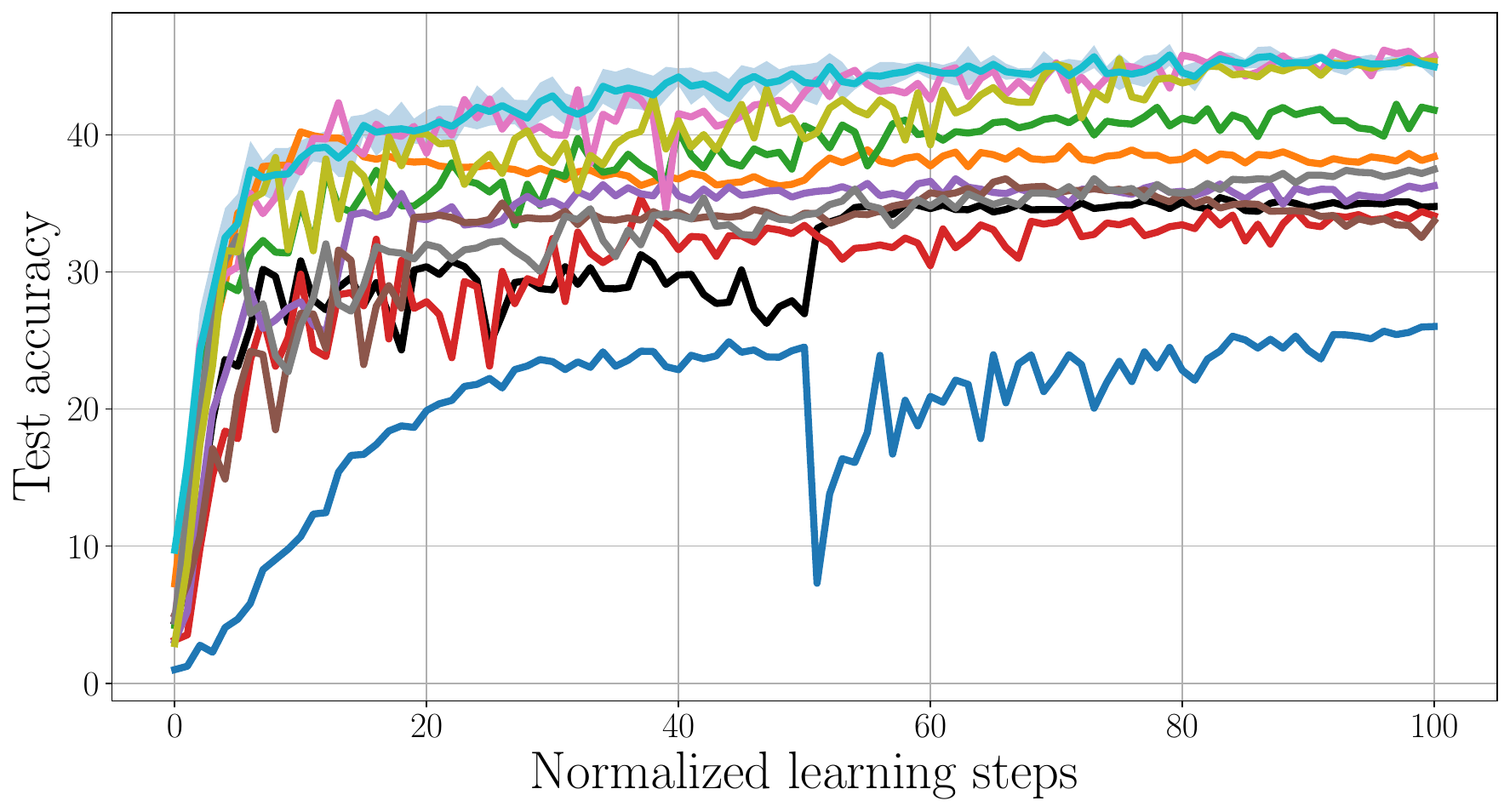}
	}
	\\
	\vspace{-2mm}
	\subfigure[60\% BadLabel]{
		\centering
		\includegraphics[scale=0.26]{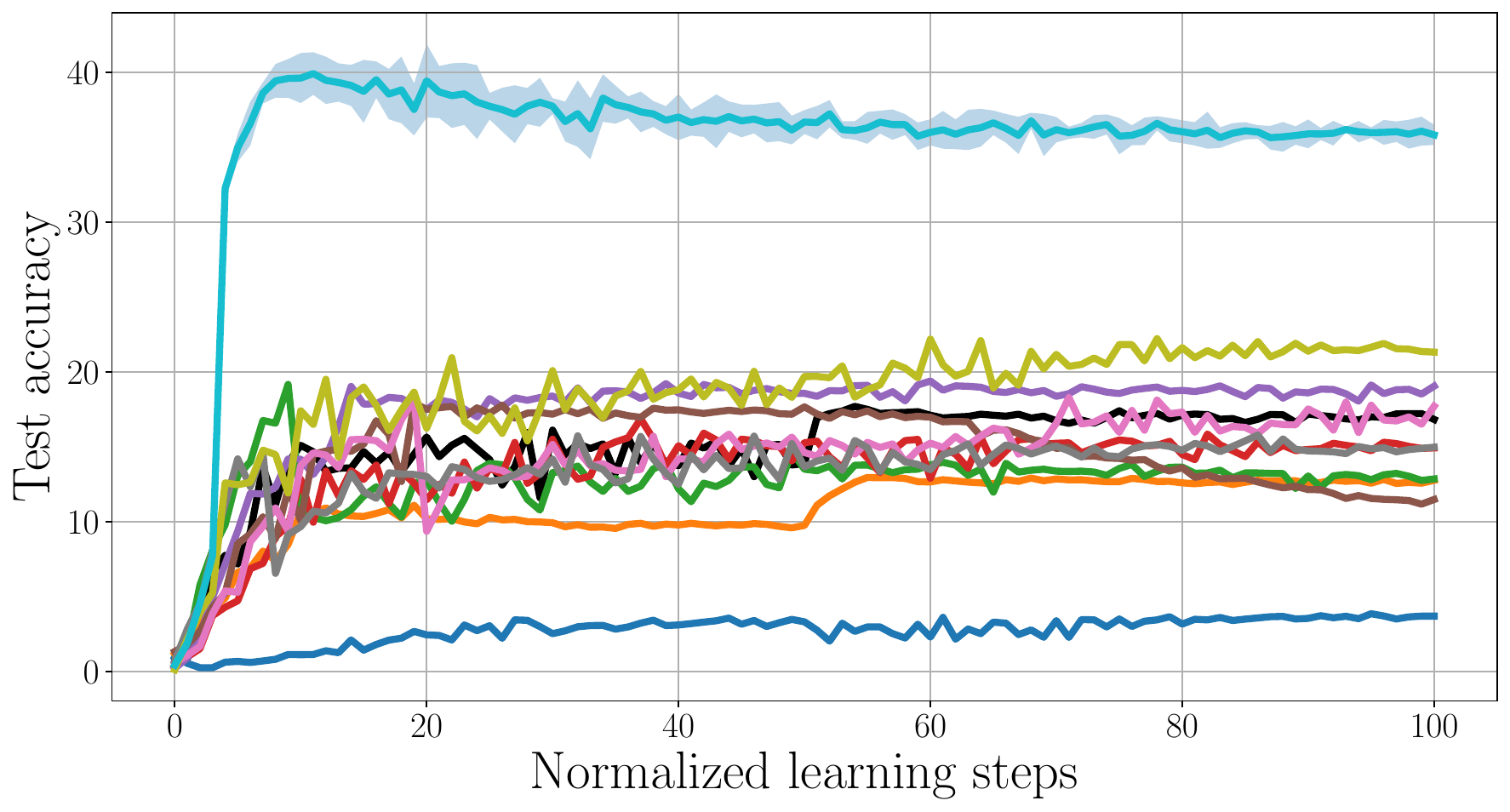}
	}
	\hspace{5mm}
	\subfigure[80\% BadLabel]{
		\centering
		\includegraphics[scale=0.26]{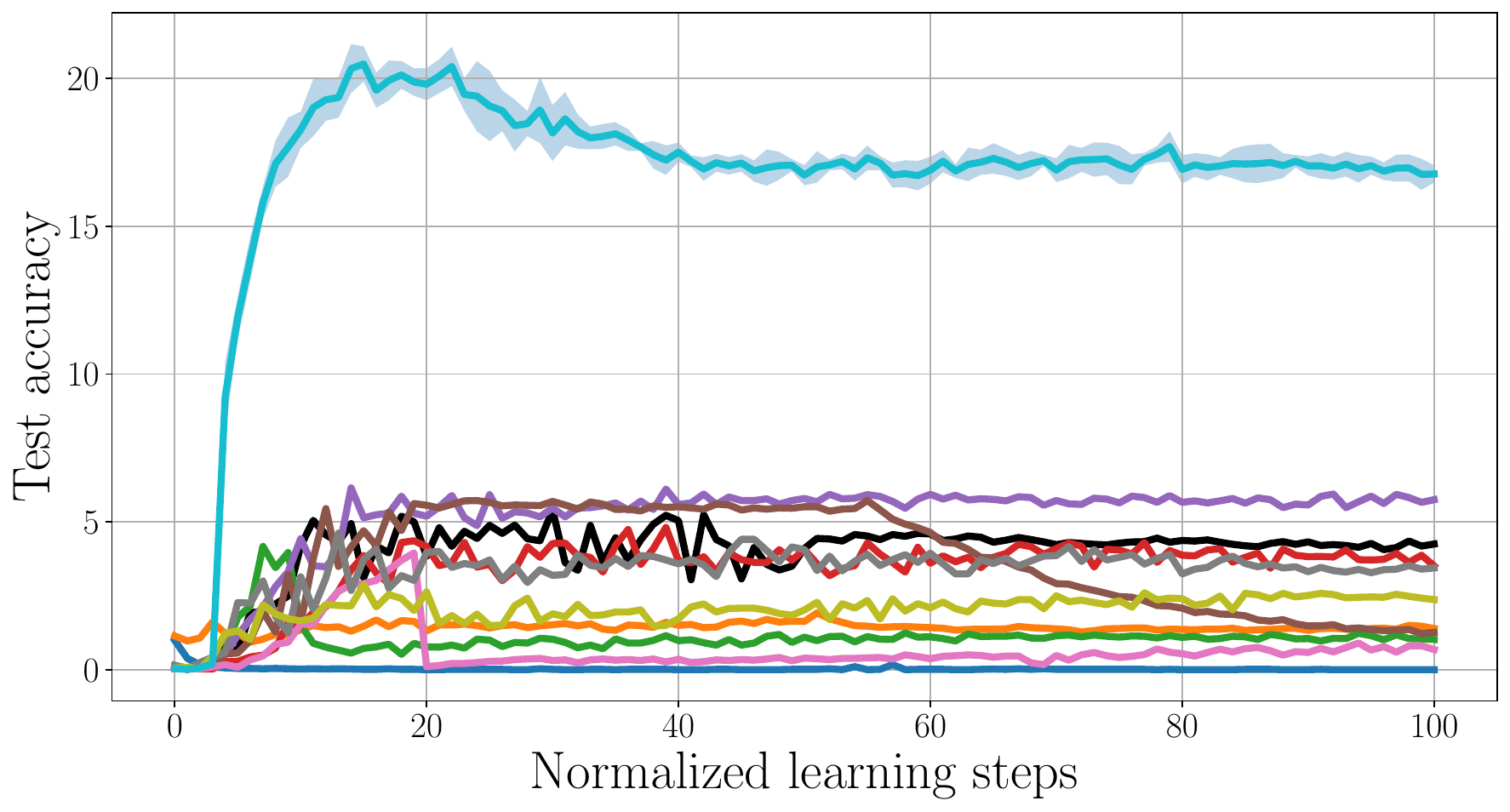}
	}

	\caption{Learning curves of several LNL algorithms on CIFAR-100 under varying BadLabel noise ratios.}
  	\vspace{-4mm}
	\label{fig:robdm_cifar100_learning_curve}
\end{figure*}

\begin{table*}[h!]
	\scriptsize
	\centering
	\renewcommand\arraystretch{1.02}
	\caption{Comparison of the test accuracy (\%) between Robust DivideMix and baseline methods on CIFAR-10 with different types and ratios of label noise. The best average performance under each noise ratio is highlighted in bold.}
	\label{exp:cifar10-robdm}
	\setlength{\tabcolsep}{2.6mm}{
		\begin{tabular}{cc|cccc|cccc|cccc}
			\toprule
			\multirow{3}*{Noise Type}&&\multicolumn{12}{c}{Method / Noise Ratio} \\
			\cmidrule{3-14} &&\multicolumn{4}{c|}{Standard Training}&\multicolumn{4}{c|}{DivideMix}&\multicolumn{4}{c}{Robust DivideMix} \\
			&&20\%&40\%&60\%&80\%&20\%&40\%&60\%&80\%&20\%&40\%&60\%&80\%\\	
			\midrule
			\multirow{2}*{Sym.}
			&Best&85.21&79.90&69.79&43.00&96.21&95.08&94.80&81.95&95.45$\pm$0.36&94.84$\pm$0.13&94.25$\pm$0.11&61.59$\pm$1.24\\
			&Last&82.55&64.79&41.43&17.20&96.04&94.74&94.56&81.58&95.28$\pm$0.38&94.71$\pm$0.16&94.11$\pm$0.12&60.98$\pm$1.21\\
			\midrule
			\multirow{2}*{Asym.}
			&Best&88.02&85.22&-&-&94.82&94.20&-&-&91.77$\pm$0.46&86.88$\pm$0.82&-&-\\
			&Last&87.28&77.04&-&-&94.46&93.50&-&-&90.62$\pm$0.38&84.02$\pm$1.65&-&-\\
			\midrule
			\multirow{2}*{IDN}
			&Best&85.42&78.93&68.97&55.34&91.97&85.84&81.59&59.06&90.44$\pm$1.09&89.71$\pm$0.74&78.12$\pm$0.31&60.64$\pm$0.46\\
			&Last&85.23&74.06&52.22&28.04&90.77&82.94&81.19&47.81&87.30$\pm$1.72&89.16$\pm$0.69&72.33$\pm$1.08&50.38$\pm$0.68\\
			\midrule
			\multirow{2}*{BadLabel}
			&Best&76.76&58.79&39.64&17.80&84.81&58.44&28.38&6.87&92.07$\pm$1.06&86.70$\pm$3.83&76.47$\pm$3.89&27.41$\pm$3.25\\
			&Last&75.31&55.72&35.66&13.44&82.13&57.65&16.21&6.12&91.76$\pm$1.27&85.96$\pm$4.33&73.29$\pm$3.81&25.20$\pm$2.72\\
                \midrule
                \multirow{2}*{Average}
                &Best&83.85&75.71&59.47&38.71&91.95&83.39&68.26&49.17&\textbf{92.43}$\pm$0.74&\textbf{89.53}$\pm$1.38&\textbf{82.95}$\pm$1.43&\textbf{49.88}$\pm$1.65\\
                &Last&82.59&67.90&43.10&19.56&90.85&82.21&63.99&45.17&\textbf{91.24}$\pm$0.93&\textbf{88.46}$\pm$1.71&\textbf{79.91}$\pm$1.67&\textbf{45.52}$\pm$1.54 \\
			\bottomrule 
	\end{tabular}}
\end{table*}

\begin{table*}[h!]
	\scriptsize
	\centering
	\renewcommand\arraystretch{1.02}
	\caption{Comparison of the test accuracy (\%) between Robust DivideMix and baseline methods on CIFAR-100 with different types and ratios of label noise. The best average performance under each noise ratio is highlighted in bold.}
	\label{exp:cifar100-robdm}
	\setlength{\tabcolsep}{2.6mm}{
		\begin{tabular}{cc|cccc|cccc|cccc}
			\toprule
			\multirow{3}*{Noise Type}&&\multicolumn{12}{c}{Method / Noise Ratio} \\
			\cmidrule{3-14} &&\multicolumn{4}{c|}{Standard Training}&\multicolumn{4}{c|}{DivideMix}&\multicolumn{4}{c}{Robust DivideMix} \\
			&&20\%&40\%&60\%&80\%&20\%&40\%&60\%&80\%&20\%&40\%&60\%&80\%\\	
			\midrule
			\multirow{2}*{Sym.}
			&Best&61.41&51.21&38.82&19.89&77.36&75.02&72.25&57.56&77.35$\pm$0.28&74.40$\pm$0.20&70.74$\pm$0.45&48.13$\pm$0.80\\
			&Last&61.17&46.27&27.01&9.27&76.87&74.66&71.91&57.08&77.06$\pm$0.28&74.16$\pm$0.23&69.93$\pm$0.59&47.84$\pm$0.82\\
			\midrule
			\multirow{2}*{IDN}
			&Best&70.06&62.48&53.21&45.77&72.79&67.82&61.08&51.50&73.49$\pm$0.28&69.47$\pm$0.18&63.64$\pm$0.21&52.74$\pm$0.73\\
			&Last&69.94&62.32&52.55&40.45&72.50&67.37&60.55&47.86&73.10$\pm$0.20&68.88$\pm$0.13&61.03$\pm$0.31&46.84$\pm$0.17\\
			\midrule
			\multirow{2}*{BadLabel}
			&Best&56.75&35.42&17.70&6.03&65.55&42.72&19.17&4.67&65.29$\pm$0.76&46.64$\pm$0.48&41.80$\pm$1.19&21.48$\pm$0.39\\
			&Last&56.30&34.90&17.05&4.18&64.96&40.92&13.04&1.10&64.49$\pm$0.96&45.26$\pm$0.40&35.91$\pm$0.67&16.91$\pm$0.41\\
			\midrule
			\multirow{2}*{Average}
			&Best&62.74&49.70&36.58&23.90&71.90&61.85&50.83&37.91&\textbf{72.04}$\pm$0.44&\textbf{63.50}$\pm$0.29&\textbf{58.73}$\pm$0.62&\textbf{40.78}$\pm$0.64\\
			&Last&62.47&47.83&32.20&17.97&71.44&60.98&48.50&35.35&\textbf{71.55}$\pm$0.48&\textbf{62.77}$\pm$0.25&\textbf{55.62}$\pm$0.52&\textbf{37.20}$\pm$0.47\\
			\bottomrule 
	\end{tabular}}
 \vspace{-2mm}
\end{table*}

To maintain consistency, we utilize PreAct-ResNet18 as the backbone of all LNL algorithms. For Robust DivideMix, we use the SGD optimizer with a momentum of 0.9 and weight decay of 0.0005 and keep the batch size at 128 for a total of 300 epochs. We initialize the learning rate to be 0.02 which was then divided by factors of 10 at the 100th and 250th epoch, respectively. We show the specific hyperparameter settings of Robust DivideMix on synthetic noise in Table~\ref{exp:rob-dm-hyper-param} (in the Appendix).
For other baseline LNL algorithms, we faithfully use optimal configurations.


Figure~\ref{fig:robdm_learning_curve} and \ref{fig:robdm_cifar100_learning_curve} show the learning curves of different LNL algorithms at different BadLabel noise ratios ranging from $20\%$ to $80\%$. Among all algorithms, Robust DivideMix has achieved the highest accuracy across all ratios. In particular, when the noise ratio is higher ($\geq$ 40\%), Robust DivideMix can significantly outperform other methods, which corroborates that Robust DivideMix can effectively handle the BadLabel. 

Besides BadLabel, we show that Robust DivideMix also performs competitively on conventional synthetic label noises. 
We apply Robust DivideMix to symmetric, asymmetric, and instance-dependent label noises on CIFAR-10 and CIFAR-100, respectively. 
Tables~\ref{exp:cifar10-robdm} and \ref{exp:cifar100-robdm} show the test accuracy of Standard Training, standard DivideMix and our Robust DivideMix, under different types of label noises. 
Tables~\ref{exp:cifar10-robdm} and \ref{exp:cifar100-robdm} show that Robust DivideMix can achieve significant improvements on BadLabel while also achieving competitive performance on other types of label noise.  
Thus, the averaged accuracy of Robust DivideMix is higher than that of baseline methods, indicating that it is a more generalizable method.

Furthermore, we also validate our method on real-world noise datasets.
CIFAR-10N~\cite{wei2021learning} is a variant of CIFAR-10 with real human annotations, which contains multiple noise types (Aggregate, Random, and Worst). In our experiments, the noise type we use is ``Worst'', with the highest real noise ratio of $40.21\%$. We adopt the same training setting as on symmetric noise of CIFAR-10 and tune $N_{\mathrm{iter}}$ to 50.
Clothing1M~\cite{xiao2015learning} is a large-scale real-world dataset that contains 1 million images and corresponding annotations from the internet. We use ResNet-50~\cite{he2016deep} with ImageNet~\cite{deng2009imagenet} pre-trained weights as the backbone network. We use the SGD optimizer with a momentum of 0.9, weight decay of 0.001, and batch size of 64 to train the networks for 80 epochs (including 2 warm-up epochs). The initial learning rate is set at 0.002 and is reduced by a factor of 10 after the completion of the first 40 epochs. We set $\lambda$ as 0.2, $N_{\mathrm{iter}}$ as 50, and $\delta$ as 0.01.
We report the test accuracy with the standard deviation of Robust DivideMix on the real-world noise datasets in Table~\ref{exp:real-world-robdm}. Compared with DivideMix, Robust DivideMix still achieves strong generalization. In other words, Robust DivideMix is a more general method that can also effectively deal with real-world noises.

\begin{table}[h!]
	\small
	\centering
	\renewcommand\arraystretch{1.12}
	\caption{Test accuracy (\%) on different real-world noise datasets.}
	\label{exp:real-world-robdm}
	\setlength{\tabcolsep}{3.8mm}{
		\begin{tabular}{c|cc}
			\toprule
			Dataset / Method&DivideMix&Robust DivideMix \\	
			\midrule
			CIFAR-10N&92.56&92.70$\pm$0.20 \\
			Clothing1M&74.76&74.13$\pm$0.29 \\
			\bottomrule 
	\end{tabular}}
        \vspace{-2mm}
\end{table}

\section{Conclusion}
In this paper, we have introduced a challenging label noise called BadLabel. We have theoretically analyzed BadLabel's algorithm and empirically justified its effectiveness in significantly degrading the performance of existing LNL algorithms.
Besides, we have proposed a robust LNL algorithm, namely, Robust DivideMix, 
specifically designed to handle the challenges posed by BadLabel. 
Additionally, we have shown that Robust DivideMix is also capable of handling conventional types of label noise, providing robust performance in various scenarios.

However, there are some limitations to our work that should be acknowledged.
Firstly, the evaluation of our proposed algorithm has primarily focused on image classification tasks, and further investigations are needed to assess its performance in other domains. Secondly, although Robust DivideMix has shown promising results, there is still room for improving the efficiency of hyperparameter tuning, such as $\lambda$ in Eq.~\eqref{eq:perturb_Y}.
Future research could include developing more effective and efficient LNL algorithms to robustly handle various types of label noise, e.g., choosing effective sample separation metrics without relying on the loss distribution assumption~\cite{lu2023label}.

We have identified the potential negative impact of this work. The method of the Badlabel could be exploited by malicious attackers. From the attacker perspective, BadLabel can serve as a powerful label-flipping attack against supervised deep learning algorithms. Attacker can generate a small number of malicious label noises, but pose a significant threat to the various deep-learning-based systems, especially in the federated learning scenarios. This is particularly harmful to AI systems when utilized in critical applications such as medical analysis and autonomous driving.


\section*{Acknowledgments}
\small
Lei Liu is partially supported by National Natural Science Foundation of China (NSFC No.62220106004), Natural Science Foundation of Shandong (Shandong NSF No.ZR2021LZH006), Taishan Scholars Program.
Bo Han is partially supported by the RGC Early Career Scheme No.22200720, NSFC General Program No.62376235, Guangdong Basic and Applied Basic Research Foundation No.2022A1515011652, HKBU Faculty Niche Research Areas No.RC-FNRA-IG/22-23/SCI/04, and HKBU CSD Departmental Incentive Scheme.
Tongliang Liu is partially supported by the following Australian Research Council projects: FT220100318, DP220102121, LP220100527, LP220200949, and IC190100031.


\bibliographystyle{IEEEtran}
\bibliography{reference}
\vspace{-3mm}
\begin{IEEEbiography}[{\includegraphics[scale=0.16]{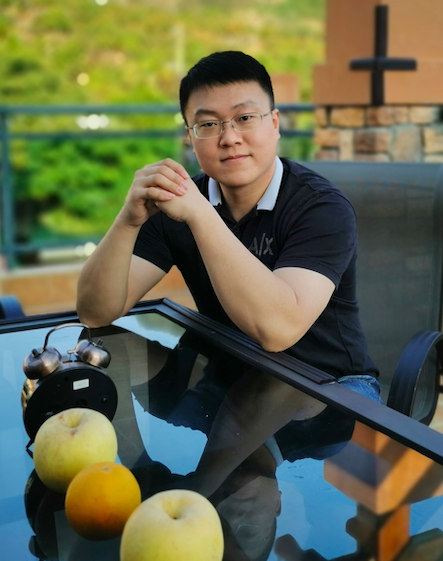}}] {Jingfeng Zhang} is a Lecturer and PhD/Doctoral Accredited Supervisor at the University of Auckland and also a Visiting Scientist at the RIKEN Center for Advanced Intelligence Project (AIP).
He obtained his Ph.D. in computer science from the National University of Singapore in 2020 and his Bachelor's degree in computer science from Shandong University's Taishan College in 2016.
He was a Postdoctoral Researcher from 2021 to 2022 and a Research Scientist in 2023 at RIKEN AIP, where he was the receiver of JST ACT-X FY2021-23, JSPS Grants-in-Aid for Scientific Research (KAKENHI) for Early-Career Scientists FY2022-23, and the RIKEN Ohbu Award 2022.
\end{IEEEbiography}
\vspace{-3mm}
\begin{IEEEbiography}[{\includegraphics[scale=0.038]{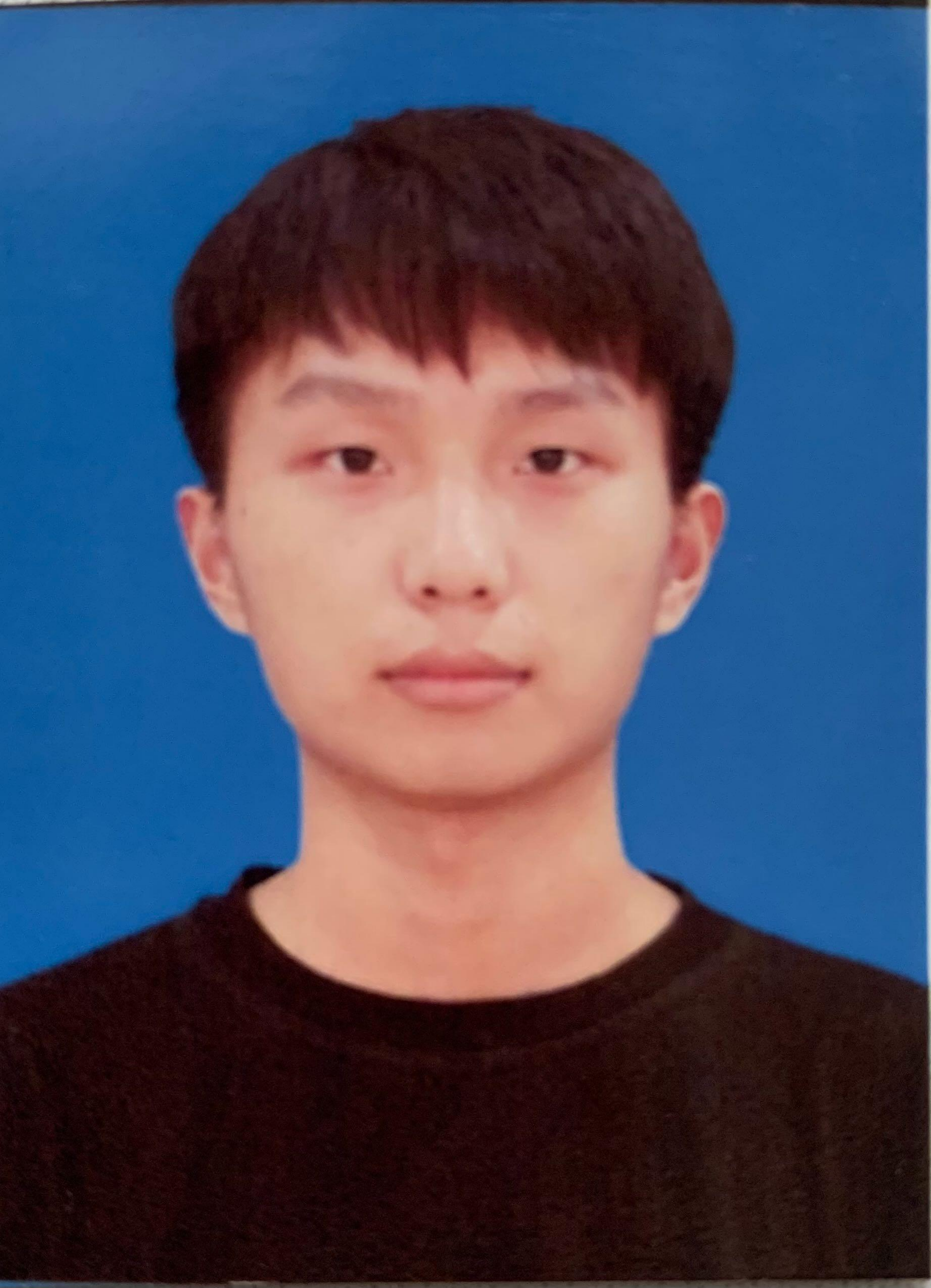}}]{Bo Song}
is currently a postgraduate student at the School of Software, Shandong University, China. He received his Bachelor's degree in software engineering from Shandong University in 2021. His research interests include adversarial attacks and label-noise learning.
\end{IEEEbiography}
\vspace{-3mm}
\begin{IEEEbiography}[{\includegraphics[scale=0.25]{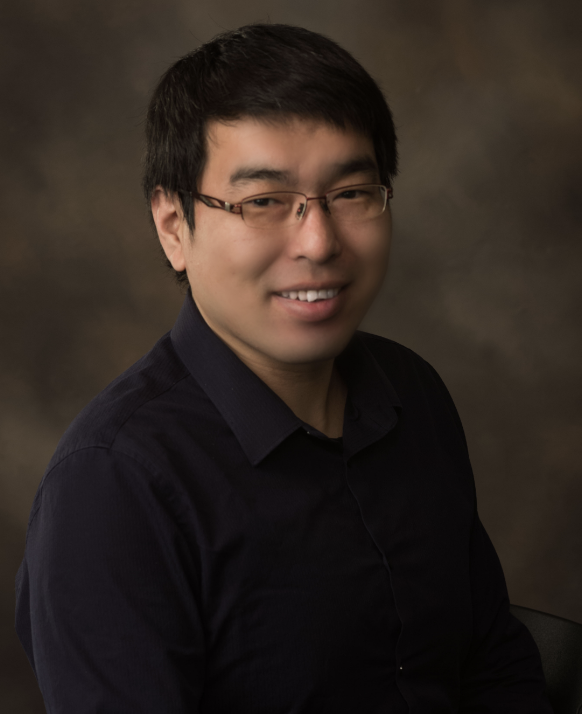}}]{Haohan Wang} is an assistant professor in the School of Information Sciences at the University of Illinois Urbana-Champaign. His research focuses on the development of trustworthy machine learning methods for computational biology and healthcare applications. He earned his PhD in computer science through the Language Technologies Institute of Carnegie Mellon University. In 2019, he was recognized as the Next Generation in Biomedicine by the Broad Institute of MIT and Harvard.
\end{IEEEbiography}
\vspace{-3mm}
\begin{IEEEbiography}[{\includegraphics[scale=0.33]{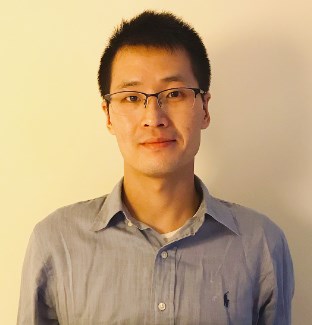}}]{Bo Han} is an Assistant Professor in Machine Learning at Hong Kong Baptist University and a BAIHO Visiting Scientist at RIKEN AIP, where his research focuses on machine learning, deep learning, foundation models and their applications. 
\end{IEEEbiography}
\vspace{-3mm}
\begin{IEEEbiography}[{\includegraphics[scale=0.33]{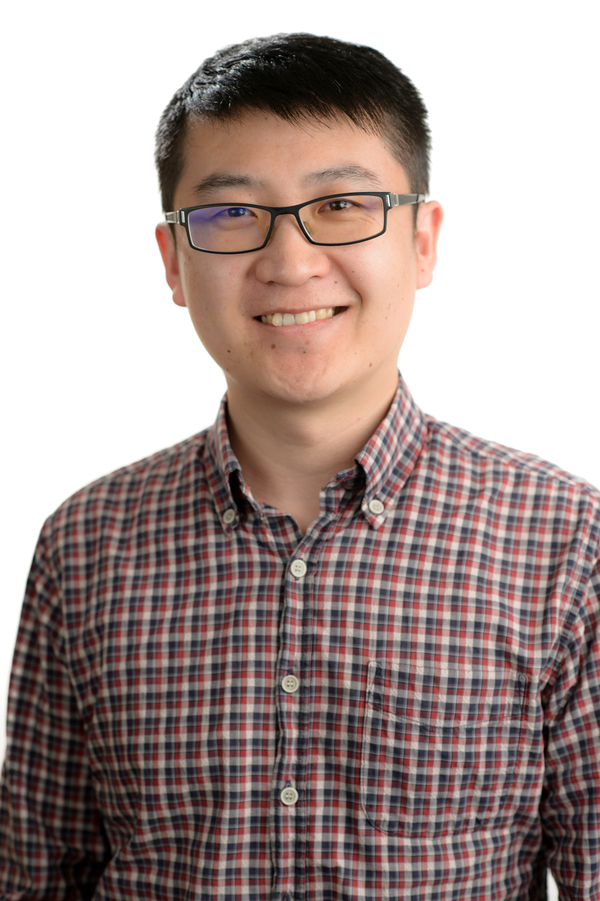}}]{Tongliang Liu} is an Associate Professor with the School of Computer Science and The Director of Sydney AI Centre at the University of Sydney. He is broadly interested in the fields of trustworthy machine learning and its interdisciplinary applications, with a particular emphasis on learning with noisy labels, adversarial learning, causal representation learning, transfer learning, unsupervised learning, and statistical deep learning theory.
\end{IEEEbiography}

\begin{IEEEbiography}[{\includegraphics[scale=0.26]{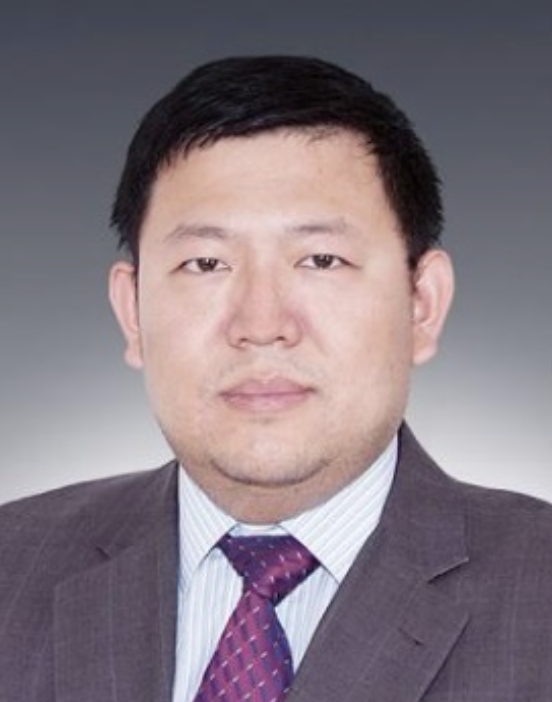}}]{Lei Liu} is a full professor in the school of software, Shandong University. He obtained the master and Ph.D degree in 2005 and 2010 from Bradford University, UK, respectively. Dr. LIU has published over 70 research papers on international conferences and journals. His research interest includes AI enabled network engineering, 5g technology, quality of service, AIoT.
\end{IEEEbiography}

\begin{IEEEbiography}[{\includegraphics[scale=0.12]{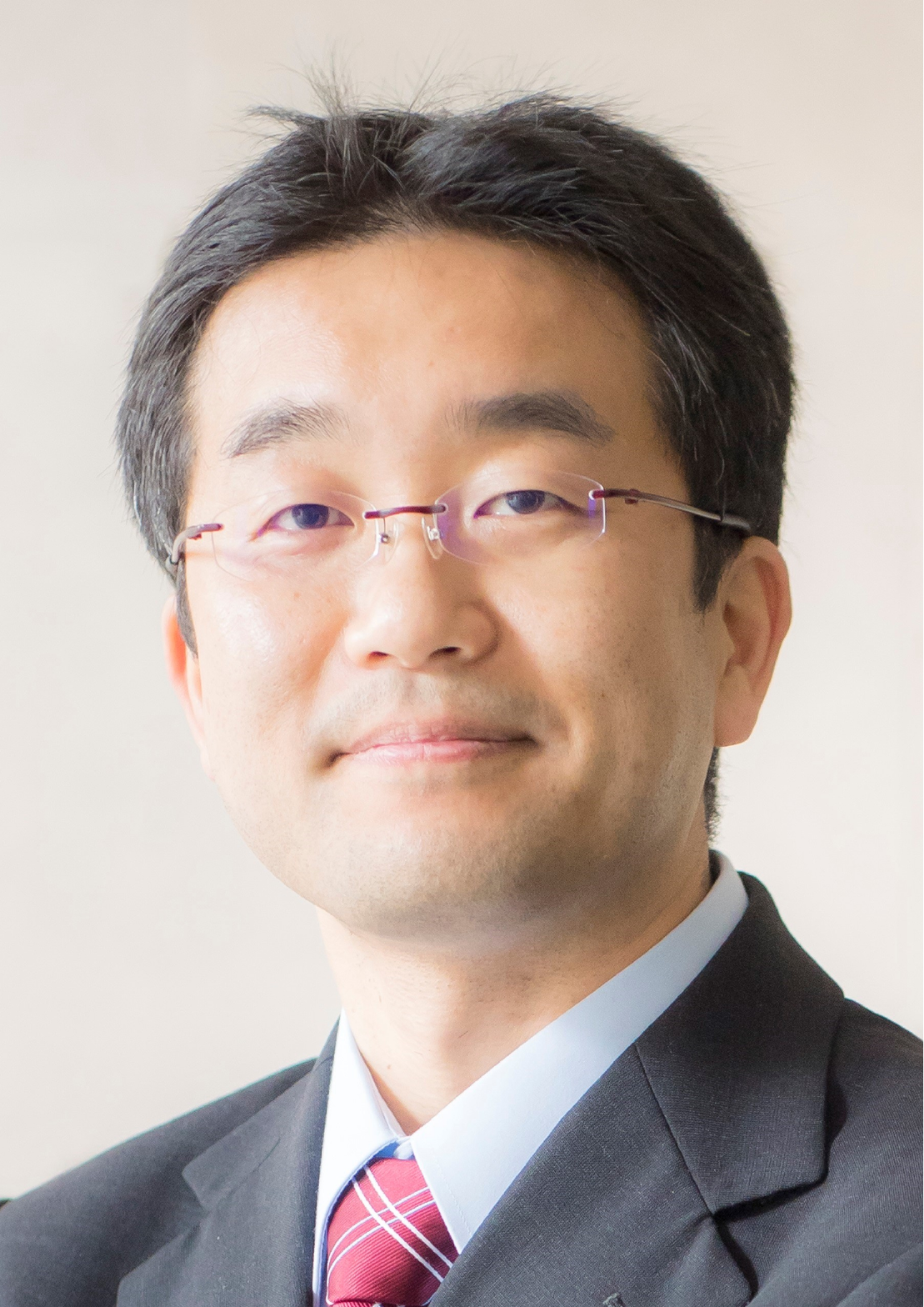}}]{Masashi Sugiyama}
received his Ph.D. in Computer Science from the Tokyo
Institute of Technology in 2001. He has been a professor at the
University of Tokyo since 2014, and also the director of the RIKEN
Center for Advanced Intelligence Project (AIP) since 2016. His
research interests include theories and algorithms of machine
learning. In 2022, he received the Award for Science and Technology
from the Japanese Minister of Education, Culture, Sports, Science and
Technology.
\end{IEEEbiography}

%
%
%
%

\newpage
\appendices
\onecolumn
\section{Mathematical Notation}
This section provides a concise reference describing the notation used throughout this paper.
\begin{table}[h!]
\centering
\caption{Notation table.}
\label{tab:notation}
\begin{tabular}{cc }
\hline
Notation & Description  \\ \hline
$f$ & A deep neural network (DNN) \\
$\theta$ & The parameter of the DNN \\
$f_{\theta}$ & The DNN $f$ parameterized by $\theta$\\
$T$ & The total iterations for generating the BadLabel, where $t \in \{1, 2,\ldots, T \} $ \\
$f_t$ & a DNN at iteration $t$ \\
$\theta$ & The parameter of a DNN $f_\theta$ \\
$D$ &  a clean $C$-class training set $D= \big \{(x_{i}, y_{i})|x_{i} \in \bR^{d}, y_{i} \in \{0,\ldots, C-1\} \big \}_{i=1}^{n}$ \\
$\rho$ & The percentage of label-flipped data over whole $n$ data \\
$D'$ &  a noisy $C$-class training set $D'= \big \{(x_{i}, y'_{i})|x_{i} \in \bR^{d}, y'_{i} \in \{0,\ldots, C-1\} \big \}_{i=1}^{n} $ \\
$\mathbbm{1}_{\{\cdot\}}(\cdot)$ &  indicator function, measuring $\rho$-distance of $D$ and $D'$, i.e., $\frac{1}{n}  \sum\limits_{i=1}^{n} \mathbbm{1}_{\{y_{i}\}}(y'_{i}) = 1-\rho$ \\
$X$ & The $n\times d$ tensor (i.e., $[x_1,\ldots, x_n]^\top$), where $x_i$ come from the training set $D$\\
$Y$ & $n\times C$ array (i.e., one-hot version of clean labels) for initializing the label optimization \\
$Y'$ & $n\times C$ array, in which $Y' \in \bR^{n \times C}$ is an optimized and noisy counterpart of $Y$ \\
$z$ & flag array $z \in \bR^{n \times C}$, whose element $z(i,j)$ denotes data $x_i$ affinity to label $j\in \{0, ..., C-1\}$\\
$\tilde{Y}$ & $n\times C$ array, in which $\tilde{Y}$ is adversarial perturbed variant of one-hot encoded noisy labels $Y'$ \\
$\lambda$ & the step size of adversarial label perturbation \\
${\rm BayesGMM}$ &  $\rm BayesGMM$ clusters per-sample losses $\{\ell(f_{\theta}(x_i),y'_i)\}^{n}_{i=1}$, where $(x_i, y'_i) \in D'$ \\
$\delta$ & The convergence value of ${\rm BayesGMM}$ \\
$N_{\mathrm{iter}}$ & The convergence iteration of ${\rm BayesGMM}$ \\
${\rm MixMatch}$ & A semi-supervised learning algorithm taking input labeled set $\cX$ and unlabeled set $\cU$ \\
$E$ & The total training epochs of ${\rm MixMatch}$, where $e \in \{1, 2,...,E\}$\\
$\theta^{e}$ & The parameter of a DNN at the epoch $e$ \\
$\theta_{k}$ & The parameter of a pair of DNNs indexed by $k$, where $k \in \{0, 1\}$ \\
$\rm {WarmUp}$ & The standard training of DNNs for a few epochs for the warm-up purposes\\
$\cX$ & A (mostly) clean labeled set \\
$\cU$ & A unlabeled set \\
$\tau_p$ & The selection threshold of adversarially perturbed labels for a clean labeled set\\
$\tau_c$ & The selection threshold of unperturbed labels for a clean labeled set\\

\hline
\end{tabular}
\end{table}

\section{Additional Experiments}
In this section, we evaluate BadLabel on various state-of-the-art LNL algorithms using different datasets and network backbones. Furthermore, we plot the learning curves of various LNL algorithms.

\subsection{Evalution of BadLabel on MNIST}
\label{Sec:cifar100-mnist-BadLabel}
For MNIST, we synthesize BadLabel noise based on the PreAct-ResNet18~\cite{he2016identity} backbone. We train the network using the SGD optimizer with a momentum of 0.5. Iteration $T$ is set to 20. The learning rate is set to 0.01, and $\alpha$ is set to 0.1.

Table~\ref{exp:mnist-BadLabel} reports the test accuracies of various methods on MNIST with different noise types and ratios. BadLabel significantly degrades the performance of multiple methods in most cases, which shows the vulnerability of existing LNL algorithms against BadLabel.

\begin{table*}[h!]
	\scriptsize
	\centering
	\renewcommand\arraystretch{1.1}
	\caption{Test accuracy (\%) on MNIST with different noise types and noise ratios. The lowest test accuracy for each method at the same noise ratio is marked in bold.}
	\label{exp:mnist-BadLabel}
	\setlength{\tabcolsep}{2.4mm}{
		\begin{tabular}{lc|cccc|cccc|cccc}
			\toprule
			\multirow{3}*{Method}&\multirow{3}*{}&\multicolumn{12}{c}{Noise Type / Noise Ratio} \\
			\cmidrule{3-14}  &&\multicolumn{4}{c|}{Sym.}&\multicolumn{4}{c|}{IDN}&\multicolumn{4}{c}{BadLabel} \\	
			&&20\%&40\%&60\%&80\%&20\%&40\%&60\%&80\%&20\%&40\%&60\%&80\% \\
			\midrule
			\multirow{2}*{Standard Training}
			&Best&98.68&97.47&97.05&77.65&93.27&77.08&53.78&34.49&\textbf{87.75}&\textbf{74.37}&\textbf{45.66}&\textbf{23.87}\\
			&Last&94.29&80.32&51.78&22.29&87.72&70.86&47.70&23.55&\textbf{82.53}&\textbf{61.31}&\textbf{39.01}&\textbf{15.93}\\
			\midrule
			Co-teaching&Best&99.19&98.96&98.73&77.30&93.91&83.84&63.26&30.07&\textbf{90.04}&\textbf{67.44}&\textbf{42.88}&\textbf{11.59}\\
			Han et al. (2018)~\cite{han2018co}&Last&97.28&94.88&92.09&70.10&91.92&74.40&57.73&28.05&\textbf{87.37}&\textbf{60.01}&\textbf{11.33}&\textbf{10.13}\\
			\midrule
			T-Revision&Best&99.24&99.06&98.56&96.24&90.90&78.82&58.58&\textbf{11.49}&\textbf{85.34}&\textbf{69.27}&\textbf{45.48}&21.83\\
			Xia et al. (2019)~\cite{xia2019anchor}&Last&99.15&99.02&98.44&96.14&87.74&69.92&46.17&\textbf{11.35}&\textbf{81.99}&\textbf{60.24}&\textbf{38.26}&16.48\\
			\midrule
			RoG&Best&-&-&-&-&-&-&-&-&-&-&-&-\\
			Lee et al. (2019)~\cite{lee2019robust}&Last&95.87&83.08&56.65&21.80&88.92&71.80&53.72&25.80&\textbf{85.62}&\textbf{65.98}&\textbf{40.58}&\textbf{18.12}\\
			\midrule
			DivideMix&Best&99.53&99.40&98.52&88.05&95.74&82.61&54.11&28.05&\textbf{85.63}&\textbf{64.76}&\textbf{44.77}&\textbf{21.18}\\
			Li et al. (2019)~\cite{li2019dividemix}&Last&98.79&96.23&91.90&61.79&88.90&68.17&43.70&21.17&\textbf{83.34}&\textbf{62.04}&\textbf{42.39}&\textbf{19.70}\\
			\midrule
			AdaCorr&Best&99.01&99.01&98.34&93.70&92.22&79.46&53.14&28.04&\textbf{84.68}&\textbf{64.86}&\textbf{42.76}&\textbf{20.92}\\
			Zheng et al. (2020)~\cite{zheng2020error}&Last&93.27&77.24&49.89&23.37&87.33&67.71&44.98&22.53&\textbf{80.53}&\textbf{59.87}&\textbf{38.34}&\textbf{17.78}\\
			\midrule
			Peer Loss&Best&99.10&98.95&98.19&93.81&92.34&85.43&58.22&47.34&\textbf{88.11}&\textbf{67.34}&\textbf{45.87}&\textbf{24.05}\\
			Liu et al. (2020)~\cite{liu2020peer}&Last&92.85&76.92&50.98&21.82&87.21&65.20&44.62&21.84&\textbf{80.49}&\textbf{59.62}&\textbf{38.85}&\textbf{18.87}\\
			\midrule
			Negative LS&Best&99.14&98.79&97.90&85.98&93.90&82.84&55.74&31.78&\textbf{88.04}&\textbf{69.95}&\textbf{47.80}&\textbf{22.60}\\
			Wei et al. (2021)~\cite{wei2021understanding}&Last&99.00&98.73&97.86&85.92&83.56&77.70&49.73&23.75&\textbf{10.87}&\textbf{25.80}&\textbf{27.03}&\textbf{10.32}\\
			\midrule
			ProMix&Best&99.75&99.77&98.07&85.50&\textbf{99.14}&96.12&69.88&41.21&99.66&\textbf{69.35}&\textbf{42.80}&\textbf{28.95}\\
			Wang et al. (2022)~\cite{wang2022promix}&Last&99.67&99.74&97.76&65.21&\textbf{97.37}&92.74&61.09&30.35&99.56&\textbf{66.33}&\textbf{35.80}&\textbf{19.09}\\
			\midrule
			SOP&Best&99.21&98.56&97.76&86.30&92.68&77.37&58.00&29.21&\textbf{91.00}&\textbf{67.60}&\textbf{48.81}&\textbf{28.57}\\
			Liu et al. (2022)~\cite{liu22sop}&Last&98.65&94.05&65.03&24.48&91.39&75.97&53.29&26.88&\textbf{84.66}&\textbf{61.78}&\textbf{37.07}&\textbf{13.95}\\
			\bottomrule 
	\end{tabular}}
\end{table*}

\subsection{Evaluation of BadLabel on DenseNet}
\label{Sec:densenet-BadLabel}
In previous experiments, we used PreAct-ResNet18 as the backbone of the LNL algorithm. To confirm that BadLabel is challenging on methods based on different network architectures, we use a 40-layer DenseNet~\cite{huang2017densely} as the backbone of the LNL algorithm for experiments. Table~\ref{exp:densenet-BadLabel} shows the test accuracy of different methods using the DenseNet backbone on CIFAR-10. BadLabel still drastically reduces the performance of methods.

\begin{table*}[h!]
	\scriptsize
	\centering
	\renewcommand\arraystretch{1.1}
	\caption{Test accuracy (\%) on CIFAR-10 with DenseNet backbone. The lowest test accuracy for each method at the same noise ratio is marked in bold.}
	\label{exp:densenet-BadLabel}
	\setlength{\tabcolsep}{2mm}{
		\begin{tabular}{lc|cccc|cc|cccc|cccc}
			\toprule
			\multirow{3}*{Method}&\multirow{3}*{}&\multicolumn{14}{c}{Noise Type / Noise Ratio} \\
			\cmidrule{3-16}  &&\multicolumn{4}{c|}{Sym.}&\multicolumn{2}{c|}{Asym.}&\multicolumn{4}{c|}{IDN}&\multicolumn{4}{c}{BadLabel} \\	
			&&20\%&40\%&60\%&80\%&20\%&40\%&20\%&40\%&60\%&80\%&20\%&40\%&60\%&80\% \\
			\midrule
			\multirow{2}*{Standard Training}
			&Best&87.71&83.42&75.23&52.33&89.57&85.75&85.59&79.00&69.69&55.16&\textbf{77.19}&\textbf{57.54}&\textbf{39.87}&\textbf{16.50}\\
			&Last&80.08&70.96&52.83&30.50&82.84&74.37&83.58&73.45&58.21&31.68&\textbf{70.71}&\textbf{51.30}&\textbf{31.86}&\textbf{12.38}\\
			\midrule
			Co-teaching&Best&86.64&85.44&80.94&53.22&87.93&70.07&83.97&76.35&58.21&43.22&\textbf{76.12}&\textbf{51.62}&\textbf{10.04}&\textbf{10.04}\\
			Han et al. (2018)~\cite{han2018co}&Last&86.51&85.22&80.69&53.11&87.82&69.60&83.80&76.06&58.02&28.00&\textbf{75.06}&\textbf{50.58}&\textbf{6.88}&\textbf{4.64}\\
			\midrule
			T-Revision&Best&76.76&74.55&58.82&48.57&78.55&77.34&82.40&71.80&66.52&54.94&\textbf{70.53}&\textbf{33.32}&\textbf{20.41}&\textbf{5.78}\\
			Xia et al. (2019)~\cite{xia2019anchor}&Last&76.38&74.27&56.92&43.20&78.16&77.23&81.64&70.68&59.50&48.24&\textbf{70.25}&\textbf{32.63}&\textbf{19.80}&\textbf{4.42}\\
			\midrule
			RoG&Best&-&-&-&-&-&-&-&-&-&-&-&-&-&-\\
			Lee et al. (2019)~\cite{lee2019robust}&Last&87.02&82.65&75.64&56.93&89.66&87.73&85.33&77.63&65.80&42.52&\textbf{82.34}&\textbf{60.24}&\textbf{31.23}&\textbf{7.78}\\
			\midrule
			DivideMix&Best&91.54&91.03&88.29&85.80&90.55&88.40&88.93&83.81&73.10&59.78&\textbf{86.88}&\textbf{57.14}&\textbf{11.68}&\textbf{5.11}\\
			Li et al. (2019)~\cite{li2019dividemix}&Last&91.14&90.62&87.94&85.30&90.02&87.72&88.03&82.99&72.19&45.84&\textbf{85.38}&\textbf{52.36}&\textbf{10.26}&\textbf{4.69}\\
			\midrule
			AdaCorr&Best&79.07&74.52&68.68&57.85&79.96&74.24&80.68&77.00&68.42&55.53&\textbf{70.90}&\textbf{48.36}&\textbf{18.27}&\textbf{3.31}\\
			Zheng et al. (2020)~\cite{zheng2020error}&Last&78.39&73.99&68.50&57.45&79.34&72.35&80.02&76.45&67.80&53.62&\textbf{69.87}&\textbf{46.96}&\textbf{14.42}&\textbf{0.75}\\
			\midrule
			Peer Loss&Best&88.68&86.56&82.92&62.99&89.09&86.57&86.02&80.28&72.57&57.03&\textbf{78.72}&\textbf{57.11}&\textbf{34.06}&\textbf{12.04}\\
			Liu et al. (2020)~\cite{liu2020peer}&Last&88.53&86.39&82.72&62.73&88.87&86.38&85.79&79.49&69.68&45.11&\textbf{76.92}&\textbf{53.87}&\textbf{32.48}&\textbf{11.08}\\
			\midrule
			Negative LS&Best&80.36&74.55&65.13&51.17&81.94&76.85&82.20&77.76&68.63&59.24&\textbf{80.36}&\textbf{57.51}&\textbf{30.83}&\textbf{10.84}\\
			Wei et al. (2021)~\cite{wei2021understanding}&Last&79.53&74.11&64.88&50.94&\textbf{73.81}&60.08&78.17&72.24&62.85&46.14&75.22&\textbf{52.55}&\textbf{25.70}&\textbf{4.78}\\
			\midrule
			PGDF&Best&92.32&91.20&89.79&80.51&91.88&85.48&89.98&86.17&78.55&58.95&\textbf{84.68}&\textbf{62.95}&\textbf{27.72}&\textbf{2.62}\\
			Chen et al. (2021)~\cite{chen2021sample}&Last&91.97&90.96&89.57&80.12&91.44&84.28&89.43&85.23&77.34&50.30&\textbf{82.33}&\textbf{58.69}&\textbf{25.20}&\textbf{2.20}\\
			\midrule
			ProMix&Best&90.44&92.28&90.85&74.92&89.89&89.14&88.99&86.59&73.51&57.06&\textbf{85.27}&\textbf{47.47}&\textbf{30.69}&\textbf{15.06}\\
			Wang et al. (2022)~\cite{wang2022promix}&Last&90.22&92.07&90.70&70.11&89.61&88.99&88.77&86.45&72.83&49.50&\textbf{84.60}&\textbf{45.86}&\textbf{25.98}&\textbf{9.68}\\
			\midrule
			SOP&Best&93.85&92.73&91.55&85.26&93.36&91.26&90.56&85.92&77.11&58.34&\textbf{87.47}&\textbf{60.88}&\textbf{23.26}&\textbf{5.06}\\
			Liu et al. (2022)~\cite{liu22sop}&Last&93.69&92.61&91.48&85.14&93.29&91.14&90.48&85.40&76.17&52.40&\textbf{87.05}&\textbf{59.67}&\textbf{17.72}&\textbf{2.69}\\
			\bottomrule 
	\end{tabular}}
\end{table*}

\subsection{Learning Curve}
In Figures~\ref{fig:cifar10_bad_learning_curve} and \ref{fig:cifar100_bad_learning_curve}, we show the learning curves of various methods on CIFAR-10 and CIFAR-100 with different noises, respectively. When combating BadLabel, methods always show lower performance throughout the learning process.

\begin{figure*}[t!]
	\centering
	\vspace{4mm}
	\subfigure[Standard Training]{
		\centering
		\includegraphics[scale=0.175]{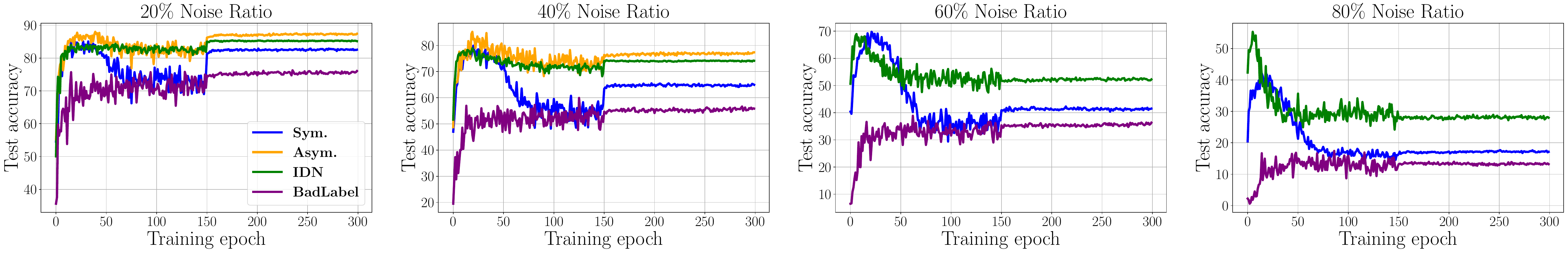}
	}\\
	\vspace{4mm}
	\subfigure[Co-teaching]{
		\centering
		\includegraphics[scale=0.175]{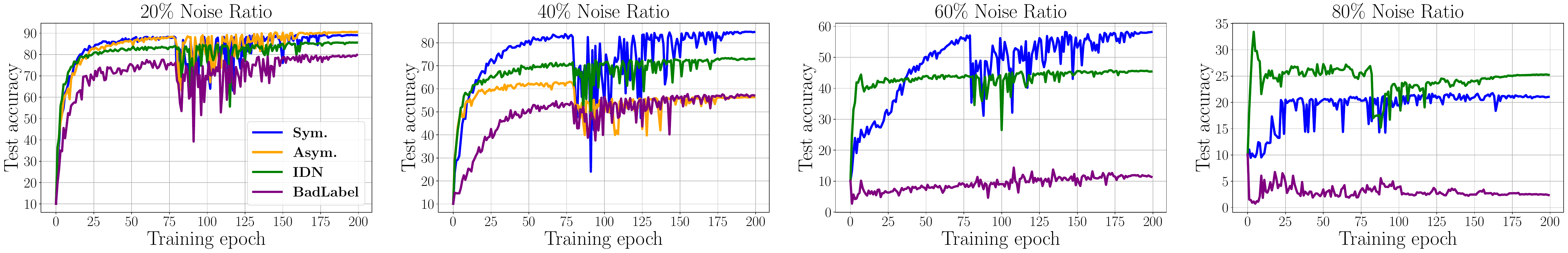}
	}\\
	\vspace{4mm}
	\subfigure[T-Revision]{
		\centering
		\includegraphics[scale=0.175]{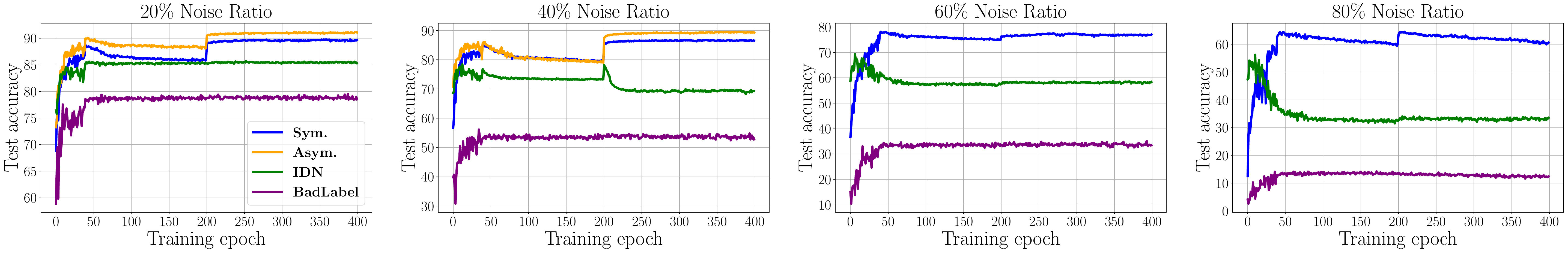}
	}\\
	\vspace{4mm}
	\subfigure[DivideMix]{
		\centering
		\includegraphics[scale=0.175]{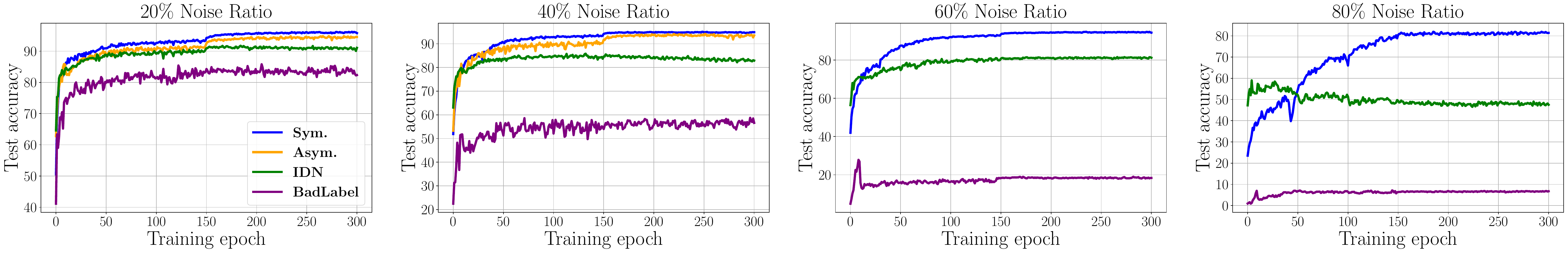}
	}\\
	\vspace{4mm}
	\subfigure[AdaCorr]{
		\centering
		\includegraphics[scale=0.175]{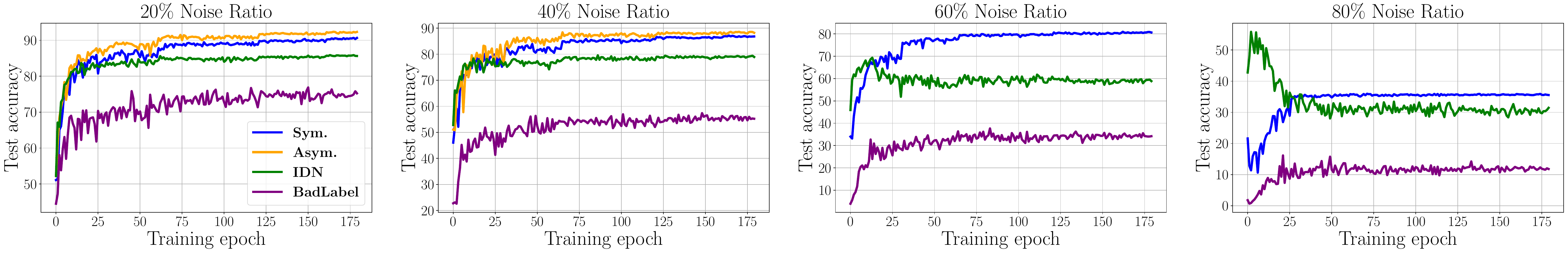}
	}\\
	\vspace{4mm}
	\subfigure[Peer Loss]{
		\centering
		\includegraphics[scale=0.175]{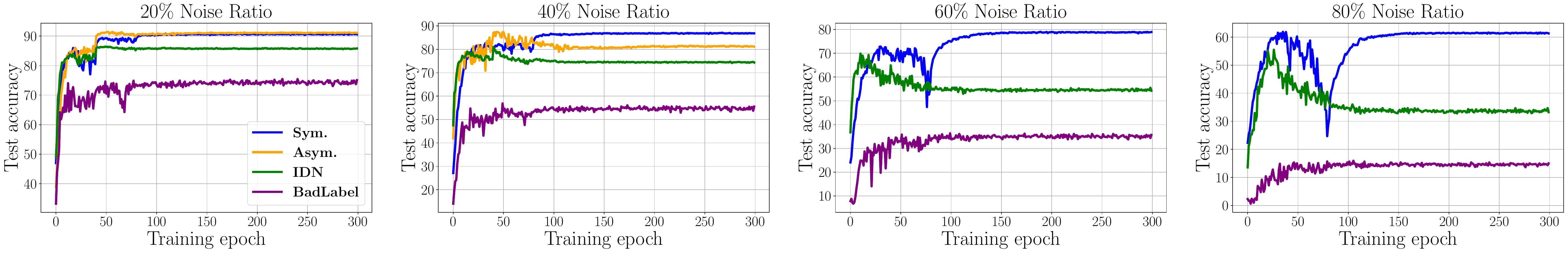}
	}\\
\end{figure*}
\begin{figure*}[t!]
	\centering
	\addtocounter{figure}{0}
	\subfigure[Negative LS]{
		\centering
		\includegraphics[scale=0.175]{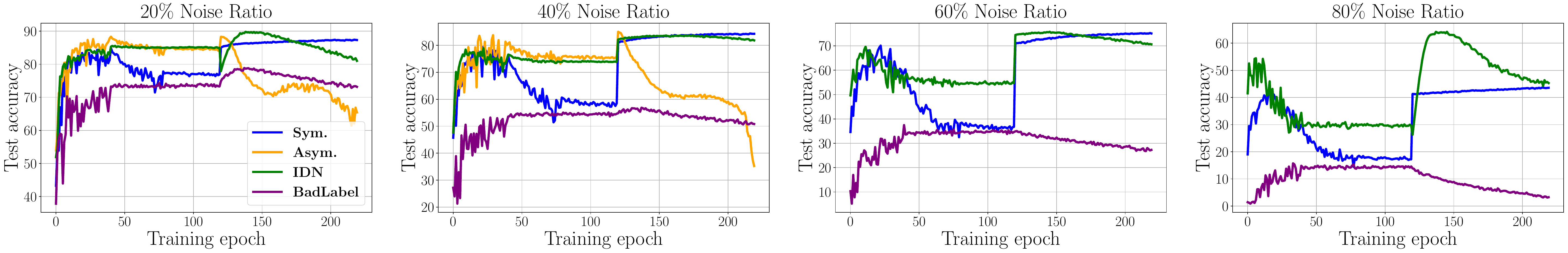}
	}\\
	\vspace{3mm}
	\subfigure[PGDF]{
		\centering
		\includegraphics[scale=0.175]{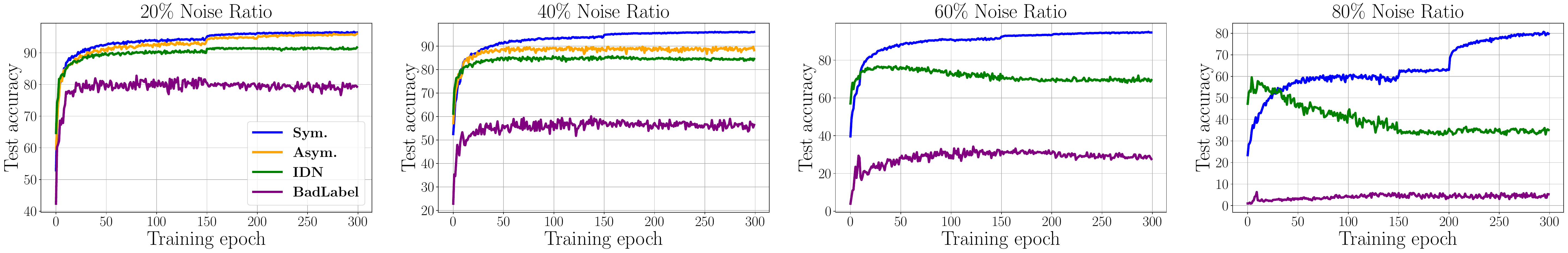}
	}\\
	\vspace{3mm}
	\subfigure[ProMix]{
		\centering
		\includegraphics[scale=0.175]{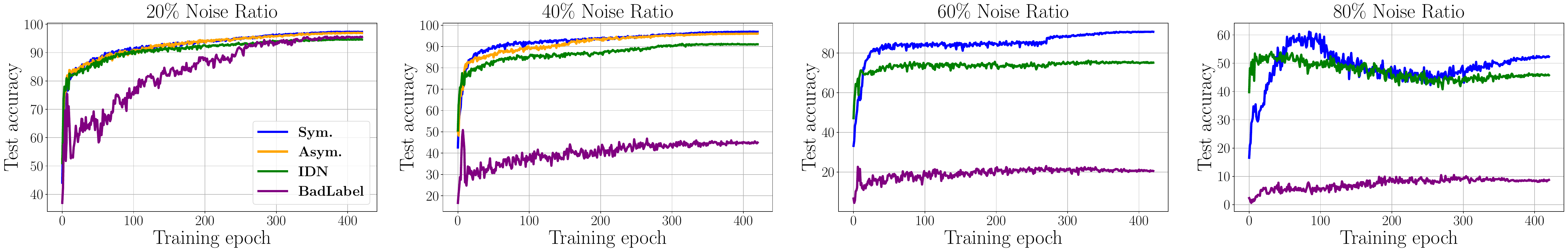}
	}\\
	\vspace{3mm}
	\subfigure[SOP]{
		\centering
		\includegraphics[scale=0.175]{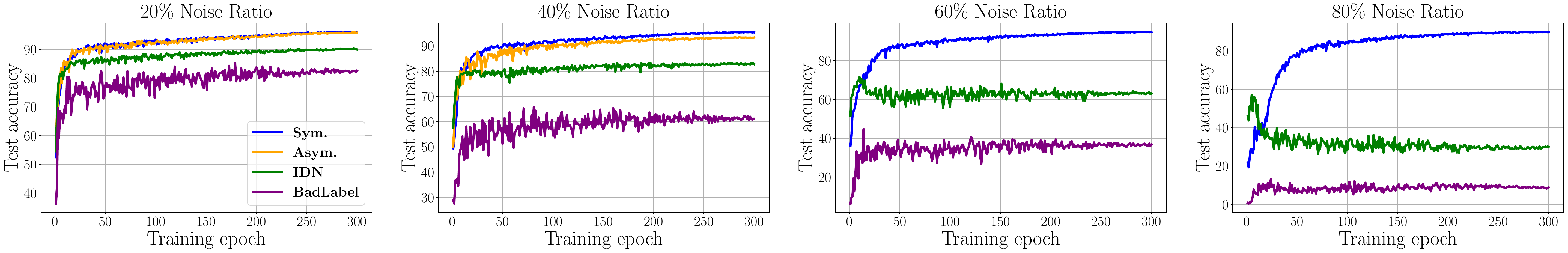}
	}\\
	\vspace{0mm}
	\caption{Learning curves of multiple LNL algorithms on CIFAR-10 with different noise types.}
	\label{fig:cifar10_bad_learning_curve}
\end{figure*}

\begin{figure*}[h!]
	\centering
	\subfigure[Standard Training]{
		\centering
		\includegraphics[scale=0.175]{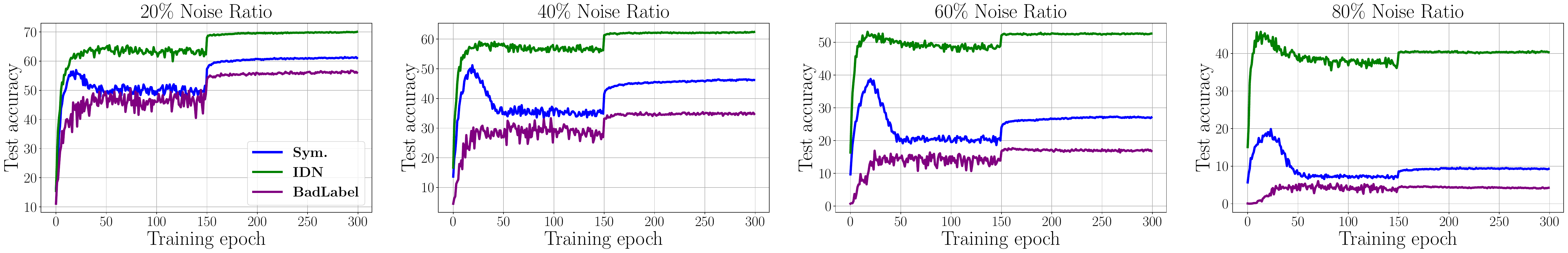}
	}\\
	\vspace{3mm}
	\subfigure[Co-teaching]{
		\centering
		\includegraphics[scale=0.175]{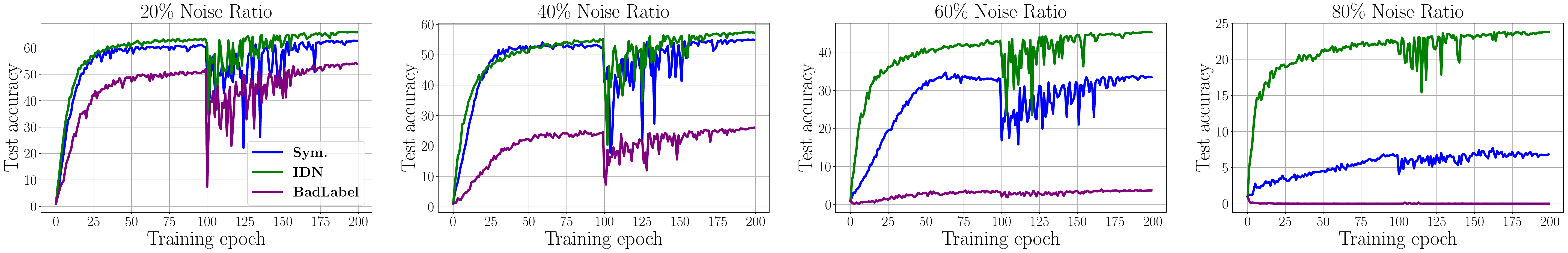}
	}\\
\end{figure*}
\begin{figure*}[h!]
	\centering
	\addtocounter{figure}{0}
	\vspace{3mm}
	\subfigure[T-Revision]{
		\centering
		\includegraphics[scale=0.175]{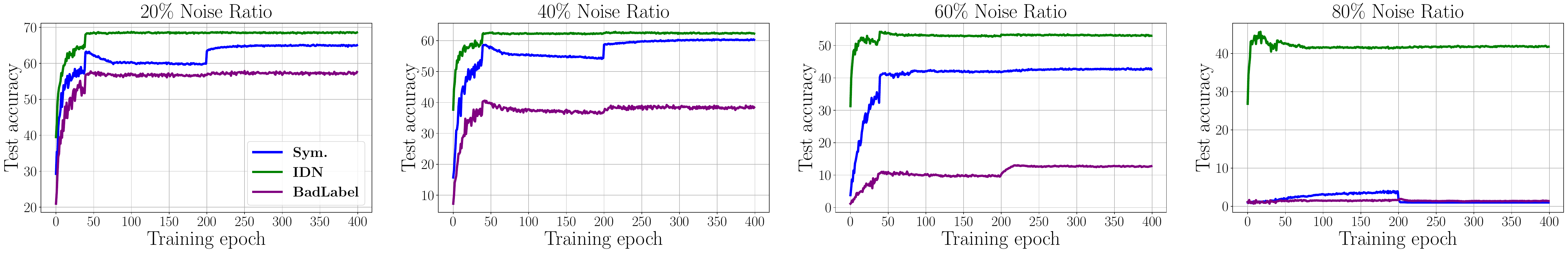}
	}\\
	\vspace{3mm}
	\subfigure[DivideMix]{
		\centering
		\includegraphics[scale=0.175]{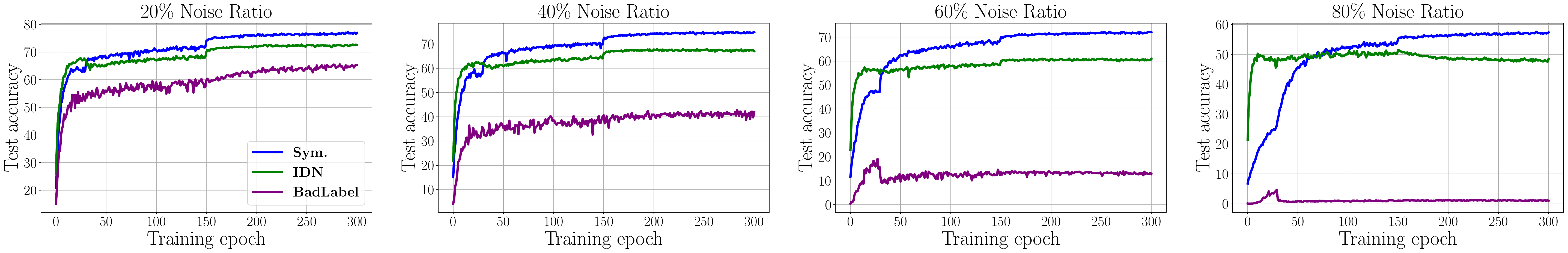}
	}\\
	\vspace{3mm}
	\subfigure[AdaCorr]{
		\centering
		\includegraphics[scale=0.175]{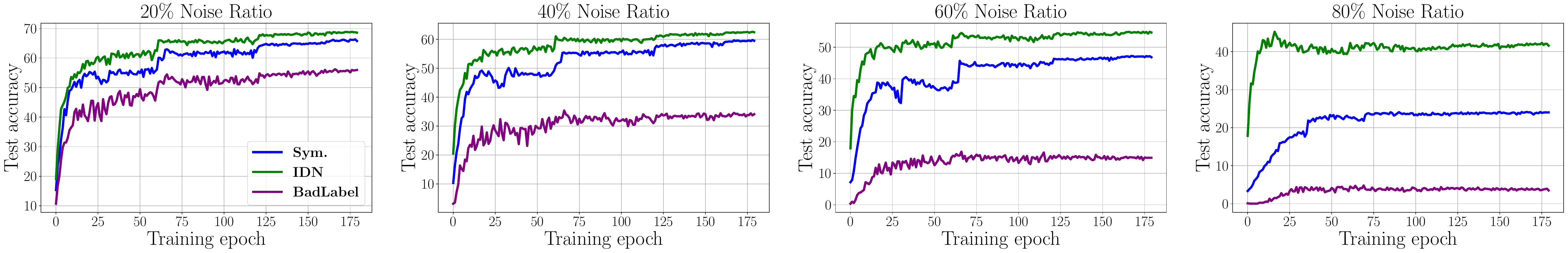}
	}\\
	\vspace{3mm}
	\subfigure[Peer Loss]{
		\centering
		\includegraphics[scale=0.175]{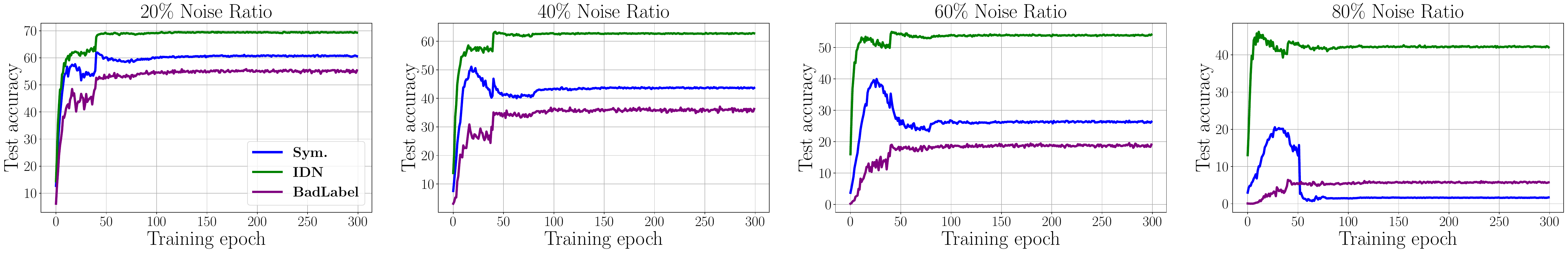}
	}\\
	\vspace{3mm}
	\subfigure[Negative LS]{
		\centering
		\includegraphics[scale=0.175]{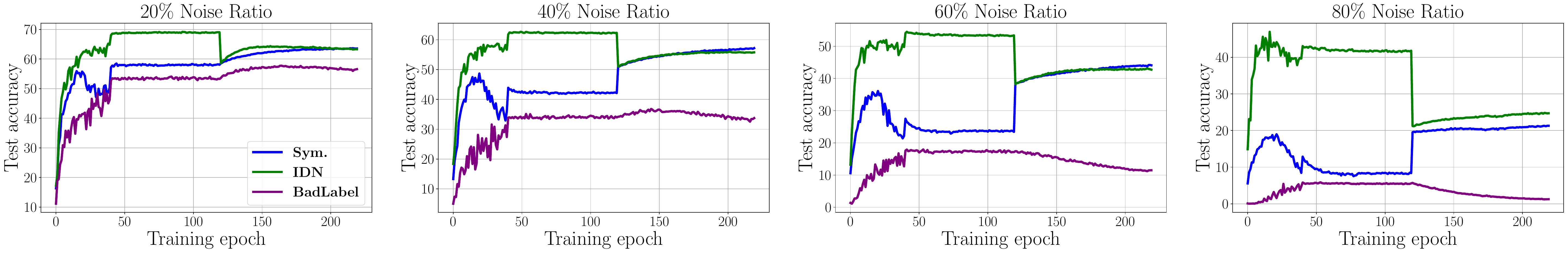}
	}\\
	\vspace{3mm}
	\subfigure[PGDF]{
		\centering
		\includegraphics[scale=0.175]{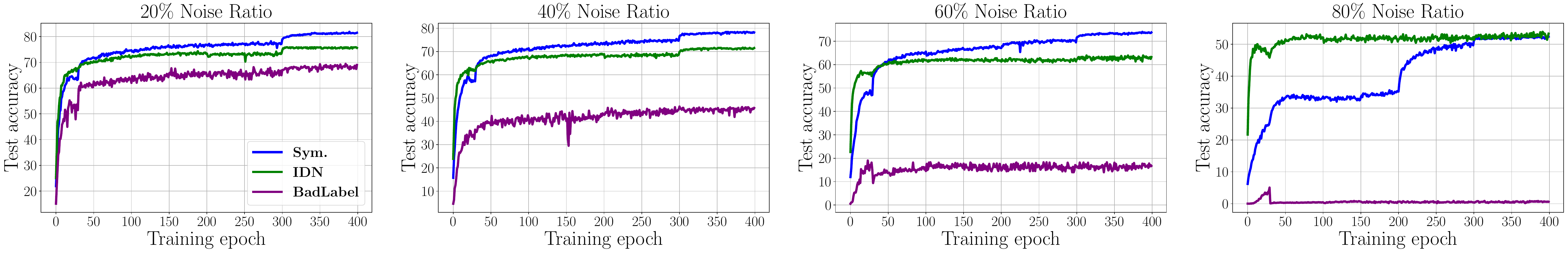}
	}\\
	\vspace{1mm}
\end{figure*}
\begin{figure*}[h!]
	\centering
	\addtocounter{figure}{0}
	\subfigure[ProMix]{
		\centering
		\includegraphics[scale=0.175]{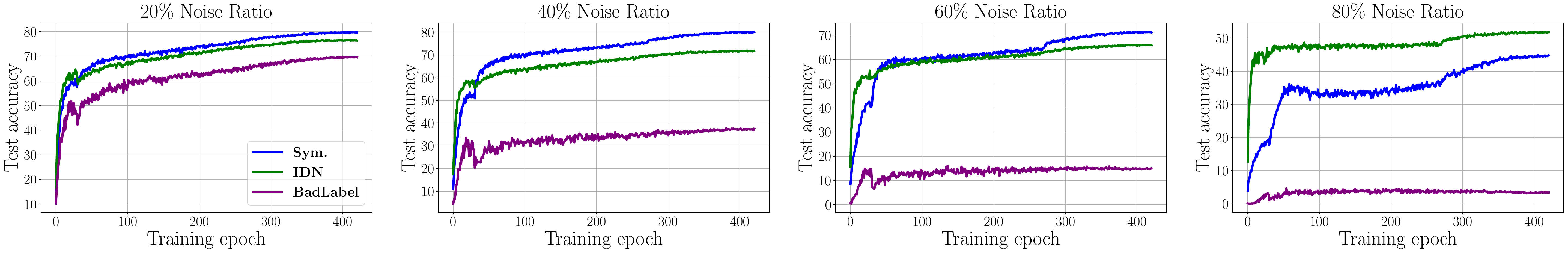}
	}\\
	\vspace{3mm}
	\subfigure[SOP]{
		\centering
		\includegraphics[scale=0.175]{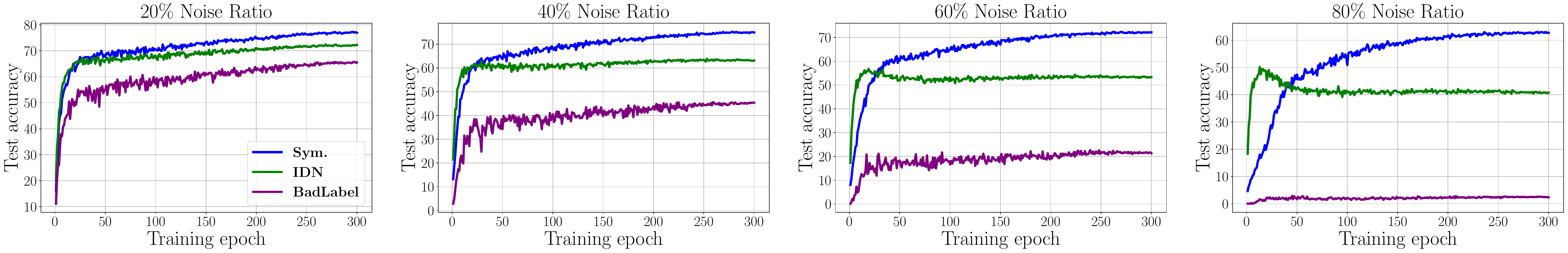}
	}\\
	
	\vspace{0mm}
	\caption{Learning curves of multiple LNL algorithms on CIFAR-100 with different noise types.}
	\label{fig:cifar100_bad_learning_curve}
\end{figure*}

\subsection{Robust DivideMix}
In this section, we provide ablation studies of Robust DivideMix.

We investigate the effects of different components in Robust DivideMix by removing the corresponding parts. In Table~\ref{exp:ablation-study}, we report the test accuracy of Robust DivideMix with different settings on CIFAR-10 and CIFAR-100 corrupted by BadLabel. 
In detail, to study the impact of introducing BayesGMM, we remove BayesGMM and replace it with the classical GMM. We present the results in the ``w/o BayesGMM'' item.
To study the effect of label perturbation, we remove the single-step perturbation by setting $\lambda$ to 0, and the results are shown in the ``w/o label perturbation'' item.
We define a division in which BayesGMM converges quickly as a high-quality division, and conversely, as a low-quality division. Robust DivideMix filters out low-quality divisions by judging whether BayesGMM converges.
To study the effect of filtering low-quality data divisions, we remove the filtering component and train the networks on new data divisions in each epoch. The results are shown in the ``w/o filtering low-quality divisions'' item.
As shown in Table~\ref{exp:ablation-study}, every component is beneficial to the good performance of Robust DivideMix. 

\begin{table*}[h!]
\vspace{4mm}
	\scriptsize
	\centering
	\renewcommand\arraystretch{1.05}
	\caption{Test accuracy (\%) of Robust DivideMix with different settings on CIFAR-10 and CIFAR-100 corrupted by BadLabel. The highest test accuracy on the same noise ratio is highlighted in bold.}
	\label{exp:ablation-study}
	\setlength{\tabcolsep}{4mm}{
		\begin{tabular}{cc|cccc|cccc}
			\toprule
			\multirow{3}*{Method}&&\multicolumn{8}{c}{Dataset / Noise Ratio} \\
			\cmidrule{3-10} &&\multicolumn{4}{c|}{CIFAR-10}&\multicolumn{4}{c}{CIFAR-100} \\
			&&20\%&40\%&60\%&80\%&20\%&40\%&60\%&80\%\\	
			\midrule
			\multirow{2}*{Robust DivideMix}
			&Best&\textbf{92.07}&\textbf{86.70}&\textbf{76.47}&\textbf{27.41}&\textbf{65.29}&\textbf{46.64}&\textbf{41.80}&\textbf{21.48}\\
			&Last&\textbf{91.76}&\textbf{85.96}&\textbf{73.29}&\textbf{25.20}&\textbf{64.49}&\textbf{45.26}&\textbf{35.91}&\textbf{16.91}\\
   			\midrule
			\multirow{2}*{w/o BayesGMM}
			&Best&88.93&74.08&44.35&15.36&62.34&43.01&30.18&6.15\\
			&Last&88.69&73.48&21.90&7.24&61.86&42.06&6.42&0.04\\
			\midrule
			\multirow{2}*{w/o label perturbation}
			&Best&89.56&60.16&67.98&21.41&64.01&38.87&39.75&20.66\\
			&Last&89.28&58.99&65.46&16.96&63.58&37.62&32.05&16.45\\
			\midrule
			\multirow{2}*{w/o filtering low-quality divisions}
			&Best&75.19&53.47&49.20&10.03&64.73&35.60&27.54&5.38\\
			&Last&75.03&51.16&12.10&5.90&63.38&34.93&3.36&0.05\\
			\bottomrule 
	\end{tabular}}
 \vspace{4mm}
\end{table*}

Table~\ref{exp:rob-dm-hyper-param} show the specific hyperparameter settings of Robust DivideMix on synthetic noise of CIFAR10/100 datasets. 

\begin{table}[h!]
\vspace{6mm}
	\scriptsize
	\centering
	\renewcommand\arraystretch{1.12}
	\caption{Hyperparameters for Robust DivideMix on CIFAR-10/100 with synthetic noise.}
	\label{exp:rob-dm-hyper-param}
	\vspace{1mm}
	\setlength{\tabcolsep}{1.8mm}{
		\begin{tabular}{lc|c|c|c|cccc}
			\toprule
			\multirow{3}*{Dataset}&\multirow{3}*{}&\multicolumn{7}{c}{Noise Type / Noise Ratio} \\
			\cmidrule{3-9}  &&\multirow{2}*{Sym.}&\multirow{2}*{Asym.}&\multirow{2}*{IDN}&\multicolumn{4}{c}{BadLabel} \\	
			&&&&&20\%&40\%&60\%&80\% \\
			\midrule
			\multirow{5}*{CIFAR-10}&warm-up&10&10&10&\multicolumn{4}{c}{4}\\
			&$N_{\mathrm{iter}}$&20&10&20&\multicolumn{4}{c}{20}\\
			&$\delta$&0.01&0.001&0.01&\multicolumn{4}{c}{0.01}\\
			&$\lambda$&0.2&0.2&0.2&0.5&0.8&1.0&1.0\\     
                &$\tau_{p}$&0.5&0.5&0.5&\multicolumn{4}{c}{0.5}\\
                &$\tau_{c}$&0.5&0.5&0.5&\multicolumn{4}{c}{0.5}\\
			\midrule
			\multirow{5}*{CIFAR-100}&warm-up&30&-&30&20&20&10&10\\
			&$N_{\mathrm{iter}}$&50&-&50&50&50&10&10\\
			&$\delta$&0.01&-&0.01&\multicolumn{4}{c}{0.01}\\
			&$\lambda$&0.2&-&0.2&\multicolumn{4}{c}{1.0}\\
                &$\tau_{p}$&0.5&0.5&0.5&0.5&0.5&0.8&0.8\\
                &$\tau_{c}$&0.5&0.5&0.5&0.5&0.5&0.8&0.8\\
			\bottomrule 
	\end{tabular}}
\end{table}

Furthermore, for the crucial hyperparameters $\lambda$, $N_{iter}$ and $\delta$, we have provided specific explanations for their selection as follows: 
For $\lambda$ in Eq.~\eqref{eq:perturb_Y}, it represents the step size of label perturbation. Overly small $\lambda$ may not be sufficient to divide the loss distribution, and overly large $\lambda$ can cause loss values to get increase significantly. We need to select $\lambda$ carefully. 
The $N_{iter}$ and $\delta$ in Eq.~\eqref{eq:BayesGMM} specify the criteria for judging the convergence of BayesGMM. $N_{iter}$ represents the number of iterations for BayesGMM, and $\delta$ represents the convergence threshold. Therefore, smaller values of $N_{iter}$ and $\delta$ imply that BayesGMM is more difficult to converge. When dealing with BadLabel, we tend to use smaller values of $N_{iter}$ and $\delta$ to filter out low-quality divisions more rigorously.

To illustrate our hyperparameter choices, we have reported relevant ablation studies, as shown in Table~\ref{exp:ablation-hyperparams}. We conducted experiments by fixing other parameters and adjusting a specific parameter. Then, we presented the average test accuracy of conventional noise at various noise ratios, as well as the test accuracy of BadLabel at different noise ratios.

\begin{table}[h!]
\vspace{4mm}
	\scriptsize
	\centering
	\renewcommand\arraystretch{1.05}
	\caption{Ablation studies on hyperparameters $\lambda$, $N_{iter}$ and $\delta$. The highest accuracy in each group is highlighted in bold.}
	\label{exp:ablation-hyperparams}
	\setlength{\tabcolsep}{2.9mm}{
		\begin{tabular}{ccc|cccccc|cccc}
			\toprule
			\multirow{2}*{Dataset}&\multirow{2}*{Noise}&&\multicolumn{6}{c|}{$\lambda$}&\multicolumn{4}{c}{$N_{iter}$ \& $\delta$} \\
			\cmidrule{4-13} 
			&&&0&0.2&0.5&0.8&1.0&1.5&10\&0.001&10\&0.01&20\&0.01&50\&0.01\\	
			\midrule
			\multirow{14}*{CIFAR-10}
			&\multirow{2}*{Sym.}&Best&85.44&\textbf{86.53}&81.28&74.46&70.10&64.22&84.75&84.55&\textbf{86.53}&83.22\\
			&&Last&85.21&\textbf{86.27}&80.89&73.69&69.70&60.20&83.66&83.20&\textbf{86.27}&80.60\\ \cmidrule{2-13} 
			&\multirow{2}*{Asym.}&Best&\textbf{89.95}&89.33&80.11&62.03&58.15&48.71&\textbf{89.33}&88.85&83.14&83.15\\
			&&Last&87.17&\textbf{87.32}&79.50&60.42&55.44&43.31&\textbf{87.32}&86.47&82.59&82.01\\ \cmidrule{2-13} 
			&\multirow{2}*{IDN}&Best&77.63&\textbf{79.73}&62.25&60.99&50.27&43.14&71.25&72.56&\textbf{79.73}&69.54\\
			&&Last&72.26&\textbf{74.79}&61.88&60.12&49.95&41.11&70.96&71.32&\textbf{74.79}&69.01\\ \cmidrule{2-13} 
			&\multirow{2}*{20\% BadLabel}&Best&89.56&87.15&\textbf{92.07}&86.79&70.52&68.44&80.31&80.19&\textbf{92.07}&76.77\\
			&&Last&89.28&86.87&\textbf{91.76}&86.60&70.01&67.23&79.13&79.44&\textbf{91.76}&76.48\\ \cmidrule{2-13} 
			&\multirow{2}*{40\% BadLabel}&Best&60.16&70.33&80.10&\textbf{86.70}&73.82&64.46&83.63&83.63&\textbf{86.70}&85.59\\
			&&Last&58.99&70.06&79.56&\textbf{85.96}&73.72&62.28&78.88&79.45&\textbf{85.96}&85.12\\ \cmidrule{2-13} 
			&\multirow{2}*{60\% BadLabel}&Best&67.98&45.25&52.26&50.52&\textbf{76.47}&40.95&71.50&70.91&\textbf{76.47}&74.53\\
			&&Last&65.46&45.03&51.77&50.25&\textbf{73.29}&40.58&70.82&70.11&73.29&\textbf{73.95}\\ \cmidrule{2-13} 
			&\multirow{2}*{80\% BadLabel}&Best&21.41&6.12&6.53&7.92&\textbf{27.41}&22.51&\textbf{29.39}&26.41&27.41&6.27\\
			&&Last&16.96&5.88&6.17&7.62&\textbf{25.20}&20.55&24.14&22.70&\textbf{25.20}&5.93\\
			\midrule
			\multirow{12}*{CIFAR-100}
			&\multirow{2}*{Sym.}&Best&67.15&\textbf{67.66}&65.25&58.47&56.32&50.30&47.12&60.52&65.44&\textbf{67.66}\\
			&&Last&67.01&\textbf{67.25}&64.45&56.17&53.92&44.08&40.13&58.47&62.30&\textbf{67.25}\\ \cmidrule{2-13} 
			&\multirow{2}*{IDN}&Best&\textbf{64.96}&64.84&60.50&55.53&48.71&40.33&38.05&40.11&60.02&\textbf{64.84}\\
			&&Last&\textbf{64.55}&62.46&59.04&54.34&44.77&37.52&35.58&36.37&58.54&\textbf{62.46}\\ \cmidrule{2-13} 
			&\multirow{2}*{20\% BadLabel}&Best&64.01&63.21&57.21&\textbf{65.54}&65.29&60.10&54.34&62.11&62.35&\textbf{65.29}\\
			&&Last&63.58&62.00&26.49&64.26&\textbf{64.49}&58.74&50.29&61.15&61.44&\textbf{64.49}\\ \cmidrule{2-13} 
			&\multirow{2}*{40\% BadLabel}&Best&38.87&41.88&42.24&43.35&\textbf{46.64}&40.63&32.64&35.58&40.01&\textbf{46.64}\\
			&&Last&37.62&40.90&40.61&42.33&\textbf{45.26}&35.52&31.22&33.96&39.65&\textbf{45.26}\\ \cmidrule{2-13} 
			&\multirow{2}*{60\% BadLabel}&Best&39.75&35.65&35.44&32.13&\textbf{41.80}&30.91&\textbf{42.33}&41.80&21.68&17.52\\
			&&Last&32.05&32.13&34.22&30.20&\textbf{35.91}&24.48&35.25&\textbf{35.91}&19.72&11.18\\ \cmidrule{2-13} 
			&\multirow{2}*{80\% BadLabel}&Best&20.66&13.50&16.11&16.15&\textbf{21.48}&20.12&20.45&\textbf{21.48}&6.09&7.20\\
			&&Last&16.45&11.25&15.80&14.98&\textbf{16.91}&13.85&16.10&\textbf{16.91}&5.33&5.58\\
			\bottomrule 
	\end{tabular}}
\vspace{4mm}
\end{table}

\section{Additional Illustration}
\begin{figure}[h]
	\centering
	\includegraphics[width=1.0\linewidth]{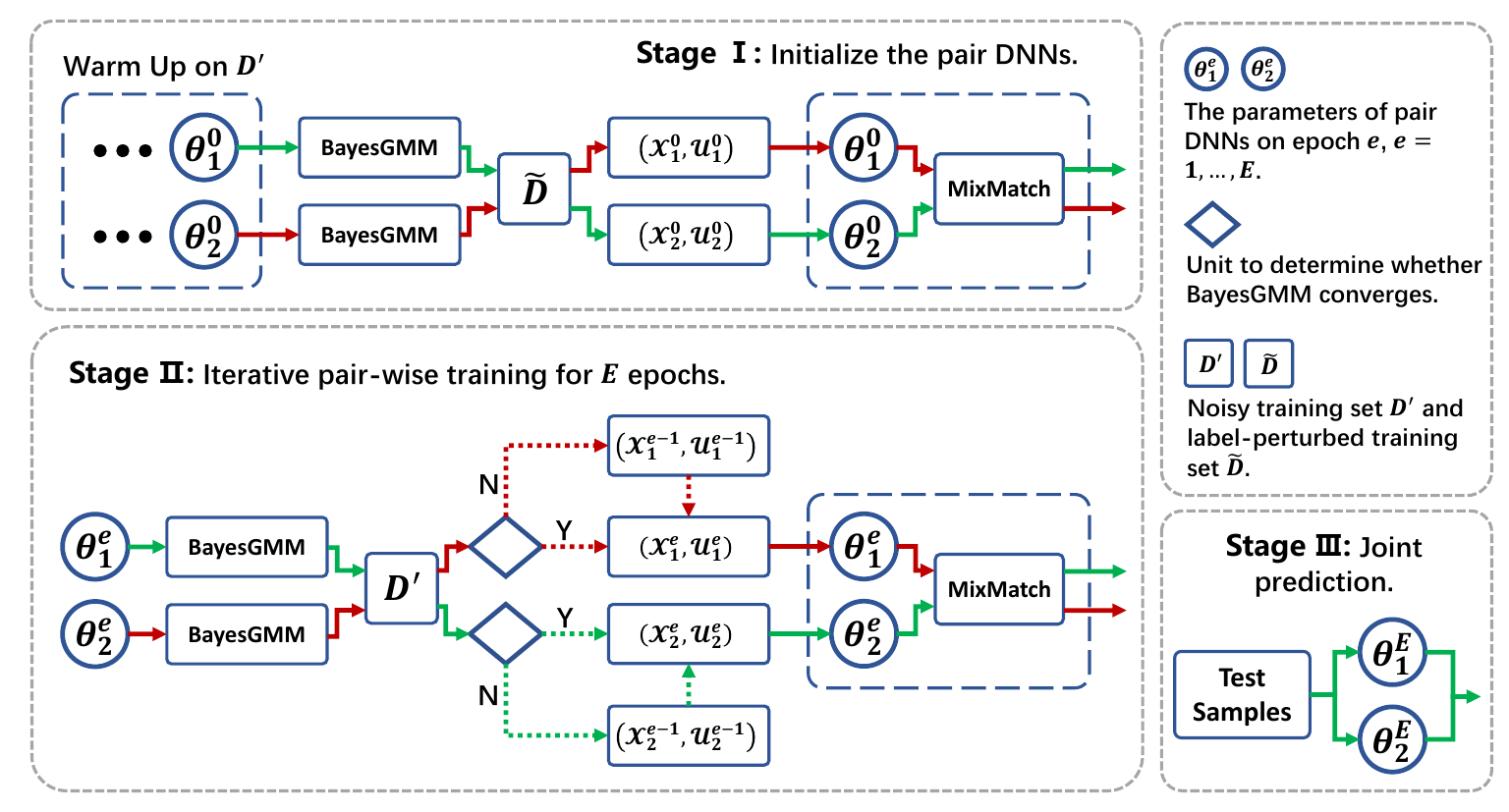}
	\vspace{-2mm}
	\caption{Illustration of Robust DivideMix. In Stage I, we first warm up pair DNNs. Then, we use BayesGMM to divide the label-perturbed dataset $\tilde{D}$ and perform a single-epoch training on it to obtain a good initialization. In Stage II, we perform pair-wise training for $E$ epochs with filtering for low-quality divisions. In Stage III, we make joint predictions based on pair DNNs.}
	\label{fig:robdm_pipeline}
\end{figure}

In this section, we present the illustration of Robust DivideMix with a three-stage workflow, as shown in Figure~\ref{fig:robdm_pipeline}.

\section{Details of Figure 1 Plotting}
In this section, we present the details of Figure~\ref{fig:prob_matrix_and_loss_dist} plotting in this paper, including transition matrices and loss distributions.

For the transition matrices, we iteratively generated multiple sets of noisy data and calculated the proportion of true labels flipping to other classes, then aggregated the results from multiple sets to obtain an empirical approximation of the flipping probability. Specifically, for each type of noise, we used different random seeds to generate 10 sets of noise and calculated the empirical label transition matrices. Finally, we mapped the probability values to the color intensity of blocks and plotted the figures.

For the loss distribution, we assume that the true labels of the samples are known and we know which training data is clean or noisy. Then, we can output the loss values associated with each training sample during the training process. Finally, we calculate the empirical probability density of clean and noisy samples. Specifically, we use the CIFAR-10 dataset with a $40\%$ noise ratio and output the loss values that are computed by DivideMix at Epoch $\#10$.

\end{document}